%% file: main_tech_report.tex
\documentclass[10pt, logo, twocolumn, copyright]{nvidiatechreport}

\input{includes/preamble}
\input{includes/colors}
\input{includes/math_commands}

\input{includes/macros}
\input{includes/notations}

\title{\oursabbr: Toward Region-level 4D Understanding via Perceptual Distillation}

\author{Chiao-An Yang$^1$, Ryo Hachiuma, Sifei Liu, Subhashree Radhakrishnan, Raymond A. Yeh$^1$, Yu-Chiang Frank Wang, Min-Hung Chen
\\
\vspace{2mm}
{\normalsize NVIDIA} \\}

\correspondingauthor{X}

\input{sections/0_abstract}

\begin{document}

\input{figures/teaser}

\input{sections/1_intro}

\input{sections/2_rel}

\input{sections/3_prelim}
\input{sections/4_app}
\input{sections/5_dataset}

\input{sections/6_exp}
\input{sections/7_conc}
\input{sections/8_ack}
\input{sections/X_supp}

\clearpage
{
  \small
  \bibliographystyle{unsrt}
  \bibliography{ref}
}

\end{document}

%% file: includes/preamble.tex
\usepackage[utf8]{inputenc} 
\usepackage[T1]{fontenc}    
\usepackage{url}            
\usepackage{booktabs}       
\usepackage{amsfonts}       
\usepackage{nicefrac}       
\usepackage{microtype}      
\usepackage{multirow}   
\usepackage{algorithmic}
\usepackage{algorithm}
\usepackage{wrapfig}
\usepackage{graphicx}
\usepackage{comment}
\usepackage{amsmath,amssymb} 
\usepackage{subcaption}

\usepackage{booktabs} 
\usepackage{amsthm}

\usepackage{multirow}
\usepackage{pmboxdraw}
\usepackage{mathtools}

\usepackage{thm-restate}
\usepackage{enumitem}

\usepackage{minted}
\usepackage{listings}
\usepackage[table]{xcolor}

\usepackage{tabularx}
\usepackage{pifont}

\usepackage{makecell}

%% file: includes/colors.tex
\usepackage{color}
\usepackage{soul}
\usepackage{xcolor}
\usepackage{wrapfig}

\definecolor{darkred}{rgb}{0.5, 0.0, 0.0}
\definecolor{lightred}{rgb}{1.0, 0.9, 0.9}
\definecolor{darkgreen}{rgb}{0.0, 0.5, 0.0}
\definecolor{lightgreen}{rgb}{0.9, 1.0, 0.9}
\definecolor{darkblue}{rgb}{0.0, 0.0, 0.5}
\definecolor{lightblue}{rgb}{0.9, 0.9, 1.0}

\definecolor{lightyellow}{rgb}{1.0, 1.0, 0.8}

\definecolor{orange}{rgb}{1.0, 0.5, 0.0}
\definecolor{cyan}{rgb}{0.7, 0.0, 0.7}
\definecolor{maroon}{rgb}{0.76, 0.13, 0.28}
\definecolor{burntorange}{rgb}{0.81, 0.33, 0}
\definecolor{tealblue}{rgb}{0.212, 0.459, 0.533}
\definecolor{skyblue}{rgb}{0.3 0.7, 1.0}
\definecolor{turquoise}{cmyk}{0.65,0,0.1,0.1}
\definecolor{purple}{rgb}{0.67, 0.28, 0.9}
\definecolor{darkpink}{rgb}{0.98,0.81,0.89}
\definecolor{lightgray}{gray}{0.88}
\definecolor{lightpurple}{rgb}{1.0, 0.9, 1.0}
\definecolor{mygreen}{rgb}{0.5, 0.7, 0.0}
\definecolor{brinkpink}{rgb}{0.98, 0.38, 0.5}

\definecolor{pp}{rgb}{0.43921569, 0.18823529, 0.62745098}
\definecolor{rr}{rgb}{0.5254902 , 0.00784314, 0.12941176}
\definecolor{bb}{rgb}{0.09019608, 0.23529412, 0.37647059}
\definecolor{yy}{rgb}{0.49803922, 0.3372549 , 0.0}
\definecolor{gg}{rgb}{0.02352941, 0.3372549 , 0.17647059}

\definecolor{ourscolor}{rgb}{0.9, 1.0, 0.75}

\ifdefined\ShowNotes
  \newcommand{\colornote}[3]{{\color{#1}\bf{#2: #3}\normalfont}}
\else
  \newcommand{\colornote}[3]{}
\fi

\newif\ifproofread

\newcommand{\oursrow}{\rowcolor{ourscolor}}

\definecolor{checkcolor}{HTML}{305AFF}
\definecolor{crosscolor}{HTML}{E62020}

%% file: includes/math_commands.tex
\usepackage{amsmath}
\usepackage{amsfonts}
\usepackage{bm}

\def\1{\bm{1}}

\def\vp{{\bm{p}}}

\def\mA{{\bm{A}}}

\def\mE{{\bm{E}}}
\def\mF{{\bm{F}}}

\def\mI{{\bm{I}}}

\def\mM{{\bm{M}}}

\def\mP{{\bm{P}}}
\def\mQ{{\bm{Q}}}

\def\mV{{\bm{V}}}

\DeclareMathAlphabet{\mathsfit}{\encodingdefault}{\sfdefault}{m}{sl}
\SetMathAlphabet{\mathsfit}{bold}{\encodingdefault}{\sfdefault}{bx}{n}
\newcommand{\tens}[1]{\bm{\mathsfit{#1}}}

\def\tD{{\tens{D}}}
\def\tE{{\tens{E}}}

\def\gL{{\mathcal{L}}}
\def\gM{{\mathcal{M}}}

\def\sR{{\mathbb{R}}}



%% file: includes/notations.tex
\newcommand{\ours}{P4D}
\newcommand{\ourbenchmark}{R4D-Bench}
\newcommand{\ourbenchmarkabbr}{R4D}
\newcommand{\oursabbr}{4D-RGPT}

\newcommand{\improve}[1]{\textcolor{checkcolor}{\footnotesize\textbf{#1}}}

\usepackage[most]{tcolorbox}

\tcbset{tightpurple/.style={
  colback=purple,
  coltext=white,
  boxrule=0pt,
  arc=2pt,
  left=1pt,
  right=1pt,
  top=1pt,
  bottom=1pt,
  boxsep=0pt,
  nobeforeafter,
  tcbox raise base,
}}

\tcbset{tightblue/.style={
  colback=blue,
  coltext=white,
  boxrule=0pt,
  arc=2pt,
  left=1pt,
  right=1pt,
  top=1pt,
  bottom=1pt,
  boxsep=0pt,
  nobeforeafter,
  tcbox raise base,
}}

\tcbset{tightgreen/.style={
  colback=mygreen,
  coltext=white,
  boxrule=0pt,
  arc=2pt,
  left=1pt,
  right=1pt,
  top=1pt,
  bottom=1pt,
  boxsep=0pt,
  nobeforeafter,
  tcbox raise base,
}}

\tcbset{lightgraybox/.style={
  colback=lightgray,
  coltext=black,
  boxrule=0pt,
  arc=2pt,
  left=1pt,
  right=1pt,
  top=1pt,
  bottom=1pt,
  boxsep=0pt,
  nobeforeafter,
  tcbox raise base,
}}

\tcbset{lightpurplebox/.style={
  colback=lightpurple,
  coltext=black,
  boxrule=0pt,
  arc=2pt,
  left=1pt,
  right=1pt,
  top=1pt,
  bottom=1pt,
  boxsep=0pt,
  nobeforeafter,
  tcbox raise base,
}}

\tcbset{bluebox/.style={
  colback=skyblue,
  coltext=white,
  boxrule=0pt,
  arc=2pt,
  left=1pt,
  right=1pt,
  top=1pt,
  bottom=1pt,
  boxsep=0pt,
  nobeforeafter,
  tcbox raise base,
}}

%% file: sections/0_abstract.tex
\begin{abstract}
Despite advances in Multimodal LLMs (MLLMs), their ability to reason over 3D structures and temporal dynamics remains limited, constrained by weak 4D perception and temporal understanding. Existing 3D and 4D Video Question Answering (VQA) benchmarks also emphasize static scenes and lack region-level prompting.
We tackle these issues by introducing:
{\bf (a)} \oursabbr, a specialized MLLM designed to capture 4D representations from video inputs with enhanced temporal perception;
{\bf (b)} Perceptual 4D Distillation (\ours), a training framework that transfers 4D representations from a frozen expert model into \oursabbr~for comprehensive 4D perception; and
{\bf (c)} \ourbenchmark, a benchmark for depth-aware dynamic scenes with region-level prompting, built via a hybrid automated and human-verified pipeline.
Our \oursabbr~achieves notable improvements on both existing 4D VQA benchmarks and the proposed \ourbenchmark~benchmark.

\noindent\textbf{Links: \href{https://www.ca-joe-yang.com/resource/projects/4D_RGPT/}{Project Page} | \href{https://github.com/NVlabs/4D-RGPT}{GitHub} | \href{https://huggingface.co/datasets/nvidia/R4D-Bench}{Dataset}}

\end{abstract}

%% file: figures/teaser.tex
\twocolumn[{%
\renewcommand\twocolumn[1][]{#1}%
\maketitle
\vspace{-0.2cm}
    \centering
    \includegraphics[width=\linewidth]{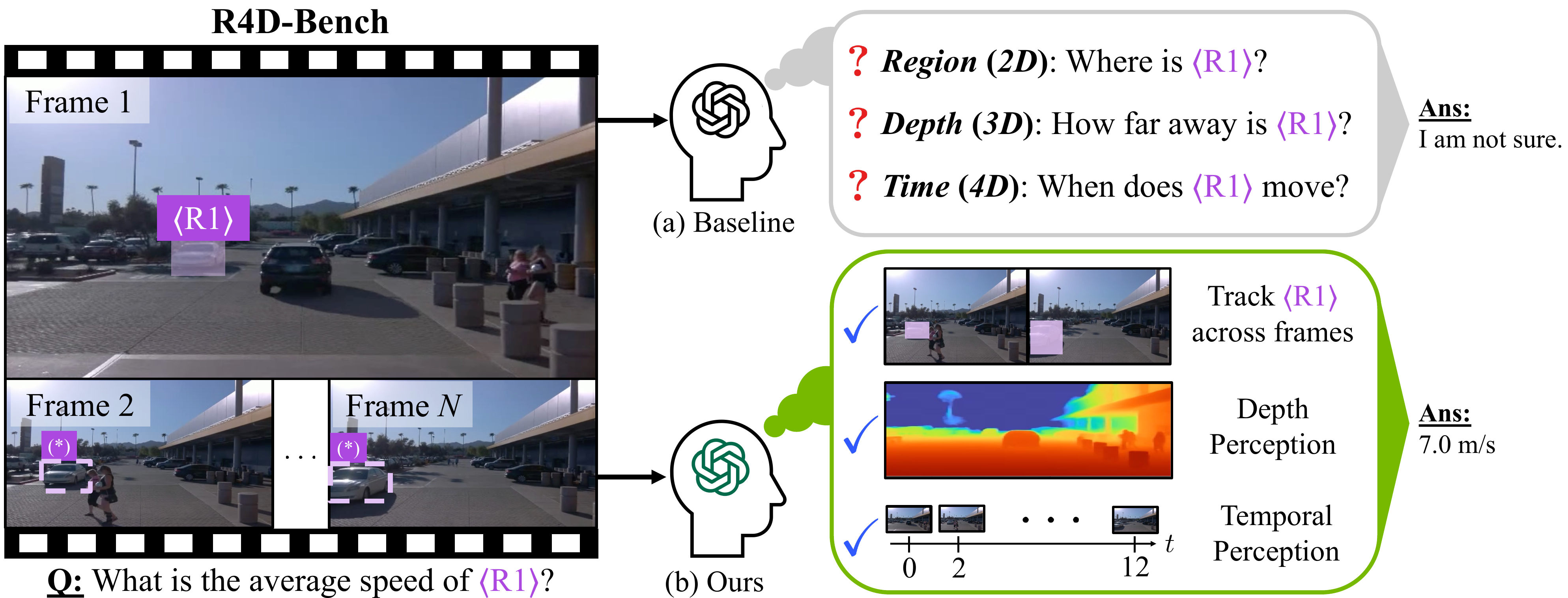}
    \vspace{-0.2cm}
    \captionof{figure}{
    \textbf{Overview of Region-level 4D Understanding.}
    4D region-level VQA, \eg, our \ourbenchmark, requires MLLMs to be able to track regions (2D), perceive depth (3D), and temporal progression (4D).
    Baseline MLLMs cannot recognize one or more of these aspects and thus fail to answer questions correctly.
    With our distillation framework, our \oursabbr~better perceives these aspects and answers accurately.
    We note that the regions labeled with \tcbox[tightpurple]{(*)} are not provided in \ourbenchmark; they are visualized for readability.
    \vspace{0.45cm}
    }
    \label{fig:teaser}
}]

%% file: sections/1_intro.tex
\section{Introduction}

By integrating visual inputs with Large Language Models (LLMs)~\cite{achiam2023gpt4,openai2025gpt5,yang2024qwen25,dubey2024llama3}, Multimodal LLMs (MLLMs) demonstrate remarkable capabilities in complex understanding across vision and language modalities. However, 
current MLLMs, even proprietary models such as GPT-4o~\cite{openai2024gpt4o}, often struggle with highly specialized tasks that require fine-grained spatial\footnote{We use ``spatial'' in this paper to refer to 3D (\ie, 2D + depth), rather than 2D as in several general video understanding works.} and temporal visual understanding.

In this paper, we advance MLLMs for one such challenging task: \textit{Region-level 4D Understanding}. This unique problem combines two critical aspects: (1) \textbf{4D understanding}, which demands answering questions regarding depth information, temporal dynamics, or object interactions in 3D space over time; and (2) \textbf{region-level understanding}, which requires grounding language queries to specific visual regions for controllable input. Region-level 4D VQA is essential for demanding real-world applications, such as autonomous driving and industrial inspection, where 4D information is critical and user queries must precisely target specific regions rather than rely on ambiguous descriptions. As an example, in~\figref{fig:teaser}, the 4D question ``{\it What is the average speed of $\textcolor{purple}{\langle R1\rangle}$?}'' specifically targets the speed of the car marked by the purple bounding box $\textcolor{purple}{\langle R1\rangle}$.

To achieve 4D understanding, previous works mainly rely on conventional Supervised Fine-Tuning (SFT)~\cite{ma2025spatialllm,zhang2025flatland,ko2025stkit,xu2025multispa} or Reinforcement Learning (RL)~\cite{wu2025vilasr,shen2025fine,ma2025spatialreasoner,ouyang2025spacer,li2025spatialladder} paradigms, optimizing primarily over the final text output using self-curated data.
However, due to the difficulty of curating large-scale, well-annotated dynamic video data, these works often struggle with dynamic scenarios.
In region-level 4D VQA, having strong 4D understanding is even more critical, as it requires tracking region movement over time.
More recently, several works~\cite{wu2025spatial,chen2025sdvlm,zheng2025vgllm,fan2025vstibench,zhou2025chat4d,cheng2025sr3d,chen2025reasoning} exploit external models to inject 3D knowledge into MLLMs to improve spatial understanding capabilities. However, external 3D knowledge mainly helps understand static videos, without fully achieving 4D understanding. Moreover, these approaches often integrate additional modules into the architecture, introducing additional inference burdens.

To address these challenges, we propose \textit{\textbf{\oursabbr}}, a specialized MLLM with effective \textit{4D perception} and thus better 4D understanding capabilities.
4D perception refers to the ability to extract low-level 4D perceptual knowledge, \eg, depth and optical flow.
Specifically, \oursabbr~perceives 4D knowledge via our proposed \textbf{P}erceptual \textbf{4}D \textbf{D}istillation (\textbf{\ours}) \textit{training-only} framework. 
\ours~adopts both latent and explicit distillation processes to effectively distill 4D perceptual knowledge from an expert 4D teacher model into the student \oursabbr.
Notably, unlike previous works, \ours~contains only \textit{training-only} modules, incurring no additional inference cost.
Finally, we introduce Timestamp Positional Encoding (TPE) to provide explicit temporal cues, enhancing MLLMs' temporal perception capability.

While various 3D/4D VQA benchmarks have been proposed recently~\cite{li2025stibench, zhou2025vlm4d, ray2024sat, jia2025omnispatial, yang2025mmsi, fan2025vstibench},
they often lack either region-prompted questions or sufficient 4D understanding challenges.
As demonstrated in~\figref{fig:teaser}, this limitation prevents comprehensive evaluation of region-based 4D VQA capabilities, namely, answering questions about specific regions (\eg, $\textcolor{purple}{\langle R1\rangle}$) in a 4D context.
To bridge this gap, we construct \textit{\textbf{\ourbenchmark}}, a new benchmark containing both static and dynamic scene understanding tasks with region-based 4D questions.

Our experiments show that \oursabbr~improves over the baseline on both non-region-based 3D/4D benchmarks ({\bf+$\bf{5.3\%}$} on average across 6 benchmarks) and our region-based \ourbenchmark~benchmark ({\bf+$\bf{4.3\%}$}),
while effectively capturing explicit 4D signals.

{\bf\noindent Our main contributions are as follows:}
\begin{itemize}
\item We propose \textit{\oursabbr} (\secref{sec:4d-rgpt}), a specialized MLLM that perceives 4D information for enhanced understanding.
\item We propose the \textit{\ours} (\secref{sec:distillation}) training framework to distill 4D perceptual knowledge into \oursabbr~without introducing additional inference cost.
\item We introduce \textit{\ourbenchmark} (\secref{sec:benchmark}), a region-based 4D VQA benchmark that requires region-level 4D understanding.
\end{itemize}

%% file: sections/2_rel.tex
\section{Related Work}

\subsection{Multimodal LLMs (MLLMs)}
\label{sec:related_mllm}
The success of LLMs~\cite{achiam2023gpt4, openai2025gpt5,touvron2023llama,touvron2023llama2,dubey2024llama3,bai2023qwen,yang2024qwen25} has inspired various MLLMs~\cite{lin2023vila, openai2024gpt4o,team2024gemini,comanici2025gemini,liu2023visual,liu2024improved,alibab2025qwen25vl,liu2025nvila} for multi-modal understanding or generation.
While several MLLMs~\cite{zhou2025strefer, liu2025vrope, shi2025causality, zeng2024timesuite, ren2024timechat} excel at video understanding,
they lack specialization in region-level or 3D/4D tasks.

\myparagraph{Region-Level MLLMs} understand specified regions within visual inputs.
Earlier works~\cite{lu2025bounding, zhao2024chatspot, pramanick2024vistallm, tian2024chatterbox, chen2023shikra, peng2024kosmos2, zhu2024minigpt4, wang2024allseeingv2, chen2024lion, lee2024collavo} use bounding box coordinates as text prompts, while others~\cite{man2025argus, lin2024draw, zhang2023gpt4roi, ma2024groma, cheng2025sr3d, wang2024asm} extract Region of Interest (RoI) visual features.
Visual markers~\cite{woo2025black, cai2024vipllava, yang2023som, lei2025scaffolding} provide intuitive region indication.
However, region-level video understanding remains challenging, especially for dynamic scenes where user queries provide sparse region annotations without temporal tracking (Fig.~\ref{fig:teaser}).
While recent works~\cite{heo2025omnirgpt, cheng2025sr3d} address this, they do not fully explore 4D dynamic scenarios.
We propose \textit{\oursabbr} (\secref{sec:4d-rgpt}) to interpret 4D spatio-temporal knowledge without 4D annotations during training.

\myparagraph{3D/4D MLLMs} focus on spatial and temporal understanding.
Previous works~\cite{zheng2025vgllm, fan2025vstibench, chen2025sdvlm, xu2025multispa,sun2025spacevista, ouyang2025spacer,li2025seetrek, cheng2025sr3d, ray2024sat, huang20253drs} enhance MLLMs with depth or 3D reconstruction models but require additional modules, introducing inference costs.
Others use SFT~\cite{ma2025spatialllm,zhang2025flatland,ko2025stkit,xu2025multispa} or RL~\cite{wu2025vilasr,shen2025fine,ma2025spatialreasoner,ouyang2025spacer,li2025spatialladder} with text-based supervision, which is insufficient for 4D perception.
We propose \textit{\ours} (\secref{sec:distillation}) to enhance 4D perception without modifying the architecture.
Prior works~\cite{radevski2023multimodal, wang2023masked} distill into vision-only encoders, \eg, ViT, VideoMAE.
3DRS~\cite{huang20253drs} employs distillation for static 3D scenes, while \ours~addresses dynamic scenes with dual distillation on latent and explicit representations to achieve 4D understanding.

\subsection{3D/4D VQA Benchmarks}
\label{sec:related_4dvqa}
Several benchmarks evaluate MLLMs' 3D and 4D understanding.
OmniSpatial~\cite{jia2025omnispatial}, VSTI-Bench~\cite{fan2025vstibench}, SAT~\cite{ray2024sat}, and MMSI-Bench~\cite{yang2025mmsi} focus on 3D spatial understanding in images.
STI-Bench~\cite{li2025stibench} is a pioneering work that introduces 4D VQA on both static and dynamic videos, while VLM4D~\cite{zhou2025vlm4d} focuses on semantic understanding in dynamic videos.
However, these benchmarks lack region-level prompting or sufficient dynamic video data (Tab.~\ref{tab:benchmark_comparison}).
We introduce \textit{\ourbenchmark} (\secref{sec:benchmark}) with region-level prompts and diverse 4D understanding tasks.

\begin{table}
    \input{tables/benchmark_comparison}
\end{table}

%% file: tables/benchmark_comparison.tex
\centering
\captionof{table}{
    \textbf{Comparison among 3D / 4D VQA Benchmarks.}
    Existing benchmarks either lack dynamic video data or region prompts,
    while our \ourbenchmark~is the first to provide both at scale.
    All benchmarks are downloaded from official sources as of August 2025,
    and the numbers of VQA might differ from the original papers. 
    Static videos contain only camera movement, while dynamic videos contain both camera and object movement.
    $^\dagger$We only adopt real-world videos from the VLM4D benchmark.
    \label{tab:benchmark_comparison}
}
\setlength{\tabcolsep}{3pt}
\resizebox{\linewidth}{!}{
\begin{tabular}{l c c c ccc}
    \specialrule{.15em}{.05em}{.05em}
    {Dataset} &
    {Regions} &
    {Input Type} &
    {FPS} &
    \# Visual & \# QA
    \\
    \midrule
    SAT-real~\cite{ray2024sat} & \ccross & Images & -
    & {196} & {150}
    \\
    MMSI-Bench~\cite{yang2025mmsi}  & \ccross & Images & - &
    {2.5k} & 1.0k
    \\
    OmniSpatial~\cite{jia2025omnispatial} & \ccross & Images & - &
    {561} & 1.5k
    \\
    VSTI-Bench~\cite{fan2025vstibench} & \ccross & Static Video & 24
    & {312} & 6k
    \\
    STI-Bench~\cite{li2025stibench}
     & \ccross & Dynamic Video & 10 $\sim$ 30 &
    369 & 2k
    \\
    VLM4D-real$^\dagger$~\cite{zhou2025vlm4d} & \ccross & Dynamic Video & 12 $\sim$ 24 &
    {600} & {1k}
    \\
    \midrule
    \oursrow \ourbenchmark~(Ours) & \ccheck & Dynamic Video & 10 $\sim$ 30 & 780 & 1.5k
    \\
    \specialrule{.15em}{.05em}{.05em}
\end{tabular}
\vspace{-0.6cm}
}

%% file: sections/3_prelim.tex
\section{Preliminaries and Notations}
\label{sec:background}

We briefly review the background and introduce notation for an MLLM and a 4D perception model.

\myparagraph{Multimodal LLMs} extend the understanding capabilities of LLMs to visual inputs such as images and videos.
The architecture typically consists of:
(a) $\tE_{\tt V}$: a vision encoder for input visuals, \eg, images or videos;
(b) $\tE_{\tt P}$: a multi-modal projector that aligns the visual and textual features within a shared space;
(c) ${\tt LLM}$: an auto-regressive model that takes in both features and generates output hidden states or tokens in a step-by-step manner;
(d) $\tD_{\tt head}$: a linear head layer that maps the hidden states to the final vocabulary space for text generation.

\myparagraph{4D Perception Models}, \eg, L4P~\cite{badki2025l4p}, encode a latent feature from input visuals for multiple 4D low-level representations.
They consist of a unified encoder $\tE_{\tt 4D}$ and specialized decoders $\tD_{m}$ for each 4D modality $m \in \gM$.
Each 4D modality $m \in \gM$ describes some per-pixel 4D properties of the input video.
For example, $m$ can be either ``depth,'' which describes the per-pixel depth values, or ``flow,'' which describes the per-pixel optical flow between adjacent frames.

We denote the input video as $\mV = [\mI^{(n)}]_{n=1:N}$ with each image frame $\mI^{(n)} \in \sR^{H \times W \times 3}$. Here, $N$ is the number of input frames and $(H, W)$ is the spatial size. Given $\mV$, we can acquire its 4D latent representation as follows,
\begin{align}
    \mF_{\tt 4D} = \tE_{\tt 4D}(\mV) \in \sR^{N' \times h' \times w' \times c'},
\label{eq:l4p}
\end{align} where $N', h', w'$ are the down-sampled number of frames, height, and width of $\tE_{\tt 4D}$'s outputs and $c'$ is the number of output channels.

For each $m$, the decoder $\tD_{m}$ decodes $\mF_{\tt 4D}$ to its corresponding low-level representation, \ie,
\begin{align}
    \mP_{m} = \tD_{m}(\mF_{\tt 4D}).
\label{eq:l4p_decode}
\end{align}
We use the following 4D modalities $\gM$ in this work:
(a) $m = {\tt depth}$ where $\mP_{\tt depth}^{(n)} \in \sR^{H \times W \times 1}$ describes the per-pixel depth values;
(b) $m = {\tt flow}$ where $\mP_{\tt flow}^{(n)} \in \sR^{H \times W \times 2}$ describes the per-pixel optical flow between adjacent frames;
(c) $m = {\tt motion}$ where $\mP_{\tt motion}^{(n)} \in \sR^{H \times W \times 1}$ describes whether a pixel is moving or static in 3D space;
(d) $m = {\tt camray}$ where $\mP_{\tt camray}^{(n)} \in \sR^{H \times W \times 6}$ describes the per-pixel Plucker ray maps.

%% file: sections/4_app.tex
\input{figures/pipeline}

\section{Approach}

\myparagraph{Overview.}
Given a video $\mV$ and a question $\mQ$, an MLLM responds with an answer $\mA$ autoregressively.
To tackle the complex, dynamic scenes presented in 4D VQA benchmarks, we develop an MLLM that can better answer questions by incorporating 4D knowledge from a teacher model and leveraging low-level representations, \eg, depth and flow, over time.
To this end, we design \textit{\textbf{\oursabbr}} to capture both \textit{latent} 4D features and \textit{explicit} 4D signals from $\mV$ with \textbf{training-only} modules.
These 4D representations enable the model to better perceive 4D knowledge during training, without introducing additional inference cost.
Additionally, to accurately capture temporal progression for answering 4D questions, we introduce Timestamp Positional Encoding (TPE) to provide explicit temporal cues to the MLLM.

To circumvent the extreme training cost and instability of training MLLMs from scratch,
we introduce our \textbf{P}erceptual \textbf{4}D \textbf{D}istillation (\textbf{\ours}) framework to distill 4D knowledge into \oursabbr~during training.
As shown in~\figref{fig:pipeline}, we leverages a frozen expert 4D perception model~\cite{badki2025l4p} (teacher), to supervise both latent and explicit 4D representations of \oursabbr~(student).
The latent distillation provides intermediate guidance on abstract 4D features, while the explicit distillation ensures accurate extraction of interpretable low-level 4D signals.
We describe the \oursabbr~architecture in~\secref{sec:4d-rgpt} and the \ours~framework in~\secref{sec:distillation}.

\subsection{4D-RGPT}
\label{sec:4d-rgpt}

Given an input video $\mV$ with $N$ sampled frames $[\mI^{(n)}]_{n=1}^N$, and the timestamps $\{t^{(n)}\}_{n=1}^N$ of each frame,
our \oursabbr~consists of training-only 4D perception modules that can extract 4D representations for distillation in \ours~(\secref{sec:distillation}).
Moreover, \oursabbr~can perceive temporal progression by incorporating timestamp positional encodings into input visual features.
In short, we use a 4D perception decoder $\tD_{\tt 4DP}$ to extract latent 4D features and prediction heads $\tD_m$ for $m\in \gM$ to extract explicit 4D signals.

\myparagraph{Latent 4D Representations.}
To capture latent 4D representations for \ours, we extract $\hat \mF_{\tt 4D}$ from the input video.
Through the video encoder $\tE_{\tt V}$, multi-modal projector $\tE_{\tt P}$, and ${\tt LLM}$, each frame $\mI^{(n)}$ is encoded as hidden state features $\mF_{\tt hidden}^{(n)} \in \sR^{h \times w \times c}$,
where $l = h w$ is the number of per-image tokens, $(h, w)$ is the spatial size of visual features, and $c$ is the hidden dimension.
We introduce a \textit{training-only} MLP as a 4D perception decoder $\tD_{\tt 4DP}$ on top of the MLLM to decode latent 4D representations $\hat \mF_{\tt 4D}^{(n)}$.
Specifically, we first sample and resize (${\tt Rearrange}$) the hidden $\mF_{\tt hidden}^{(n)}$ to match the target shape of $(N', h', w')$ in Eq.~\ref{eq:l4p}.
Thus, for each down-sampled frame $n' \in [1, N']$, we have
\begin{equation}
    \hat \mF_{\tt 4D}^{(n')} = \tD_{\tt 4DP}\left({\tt Rearrange}(\mF_{\tt hidden}^{(n)})\right).
\label{eq:latent_representations}
\end{equation}

\myparagraph{Explicit 4D Representations.}
Although $\hat \mF_{\tt 4D}$ can capture rich 4D features, explicit 4D signals, \eg, depth maps, are more interpretable and provide unambiguous supervision.
To capture explicit 4D representations for \ours, we extract explicit 4D signals $\hat \mP_{m}$ given $\hat \mF_{\tt 4D}$ via the \textit{training-only} prediction heads $\tD_{m}$ from the frozen 4D perception model.
Specifically, for each $m \in \gM$, we have
\begin{equation}
    \hat \mP_{m} = \tD_{m}(\hat \mF_{\tt 4D}).
\label{eq:explicit_representations}
\end{equation}

\myparagraph{Timestamp Positional Encoding (TPE).}
Accurate temporal perception, such as ``when'' an event occurred and ``how long'' an action took,
is fundamental to 4D VQA.
For example, to answer ``\textit{What is the average speed of the car?},'' even if the MLLM can perceive depth and knows its displacement, it still needs to understand the time duration of the video to compute speed.
Incorrect temporal perception can lead to significant errors in acquiring the displacement over the correct time duration, \ie, speed.

We observe that MLLMs struggle with temporal perception when there are no explicit time cues (see the experiments in~\secref{sec:exp_analysis} and Tab.~\ref{tab:ablation_time}).
To provide temporal cues, we encode timestamps directly into the MLLM's visual input as positional encodings.
That is, for each input frame $\mI^{(n)}$ from video $\mV$ that is sampled at time $t^{(n)}$,
we add a sinusoidal timestamp positional encoding $\vp^{(n)} \in \sR^D$ to the visual features $\tE_{\tt V}(\mI^{(n)})$ before feeding them into the $\tE_{\tt P}$, where
\begin{equation}
    \vp^{(n)}[2i] = \sin \left(\frac{t^{(n)}}{T^{\frac{2i}{D}}}\right) \text{ and } \vp^{(n)}[2i+1] = \cos\left(\frac{t^{(n)}}{T^{\frac{2i}{D}}}\right).
\label{eq:tpe}
\end{equation} Here $T$ is the maximum timescale and $i$ is the index.

\subsection{Perceptual 4D Distillation (\ours)}
\label{sec:distillation}

To answer 4D questions, MLLMs must understand not only semantic content but also various aspects of 4D knowledge, such as sub-pixel movements and numeric depth values.
For example, to answer ``\textit{Is the person moving closer to the camera?}'', the MLLM must compare the depth values of the {\it person} across frames.
Recent 3D/4D specialized MLLMs either rely on self-curated training datasets or exploit external models to enhance 3D knowledge.
However, both are insufficient for MLLMs to fully achieve 4D understanding.
Moreover, introducing external modules results in additional inference costs.
Therefore, a mechanism that provides direct supervision on the MLLM's internal 4D perception capabilities without introducing additional modules is desirable.

To this end, we propose our \ours~framework.
We leverage L4P~\cite{badki2025l4p} as the frozen expert 4D perception model (teacher) to transfer its expert representations to our student, \oursabbr.
We use the same architecture and pre-trained weights as provided in their paper.
To ensure comprehensive knowledge transfer, we propose dual-branch distillation: latent distillation and explicit distillation.

\vspace{0pt}
\myparagraph{Latent Distillation.}
We start by introducing latent distillation to supervise the MLLM's latent 4D representations, \ie, $\hat \mF_{\tt 4D}$, on the latent space.
Latent distillation serves as intermediate 4D guidance to the MLLM on the latent space.
Specifically, our latent distillation loss $\gL_{\tt LD}$ is defined to pull the margin $\Delta_{\tt LD}$ between the latent 4D features from the teacher model $\mF_{\tt 4D}$ and those from the student model $\hat \mF_{\tt 4D}$:
\begin{equation}
    \gL_{\tt LD} = \sum_{n'=1}^{N'} \Delta_{\tt LD} (\mF_{\tt 4D}^{(n')}, \hat{\mF}_{\tt 4D}^{(n')}).
    \label{eq:loss_ld}
\end{equation}

\myparagraph{Explicit Distillation.}
On the other hand, we introduce explicit distillation to supervise the MLLM's explicit 4D representations, \ie, $\hat \mP_{m}$, on the signal space.
Explicit distillation provides direct, interpretable supervision to ensure the MLLM captures accurate 4D signals in $\gM$.
Specifically, our explicit distillation loss $\gL_{\tt ED}$ is defined to pull the margin $\Delta_m$ between the explicit 4D signals from the teacher model $\mP_{m}$ and those from the student model $\hat \mP_{m}$:
\begin{equation}
    \gL_{\tt ED} = \sum_{n=1}^N \sum_{m \in \gM} \lambda_m \Delta_m (\mP_{m}^{(n)}, \hat{\mP}_{m}^{(n)}),
    \label{eq:loss_ed}
\end{equation} where $\lambda_m$ describes the loss weights of each $m$.

\myparagraph{Training.}
We optimize our \oursabbr~using both SFT and \ours.
The overall loss function is a combination of the standard cross-entropy SFT loss $\gL_{\tt SFT}$, latent distillation loss $\gL_{\tt LD}$, and explicit distillation loss $\gL_{\tt ED}$.
We train on various 3D / 4D conversation datasets, including RoboFAC~\cite{lu2025robofac}, SAT~\cite{ray2024sat}, VSTI-Bench~\cite{fan2025vstibench} (the training split), and Wolf~\cite{li2024wolf}.
Please refer to the supplementary material for more training details.

%% file: figures/pipeline.tex
\begin{figure*}[t]
    \centering
    \includegraphics[width=0.99\textwidth]{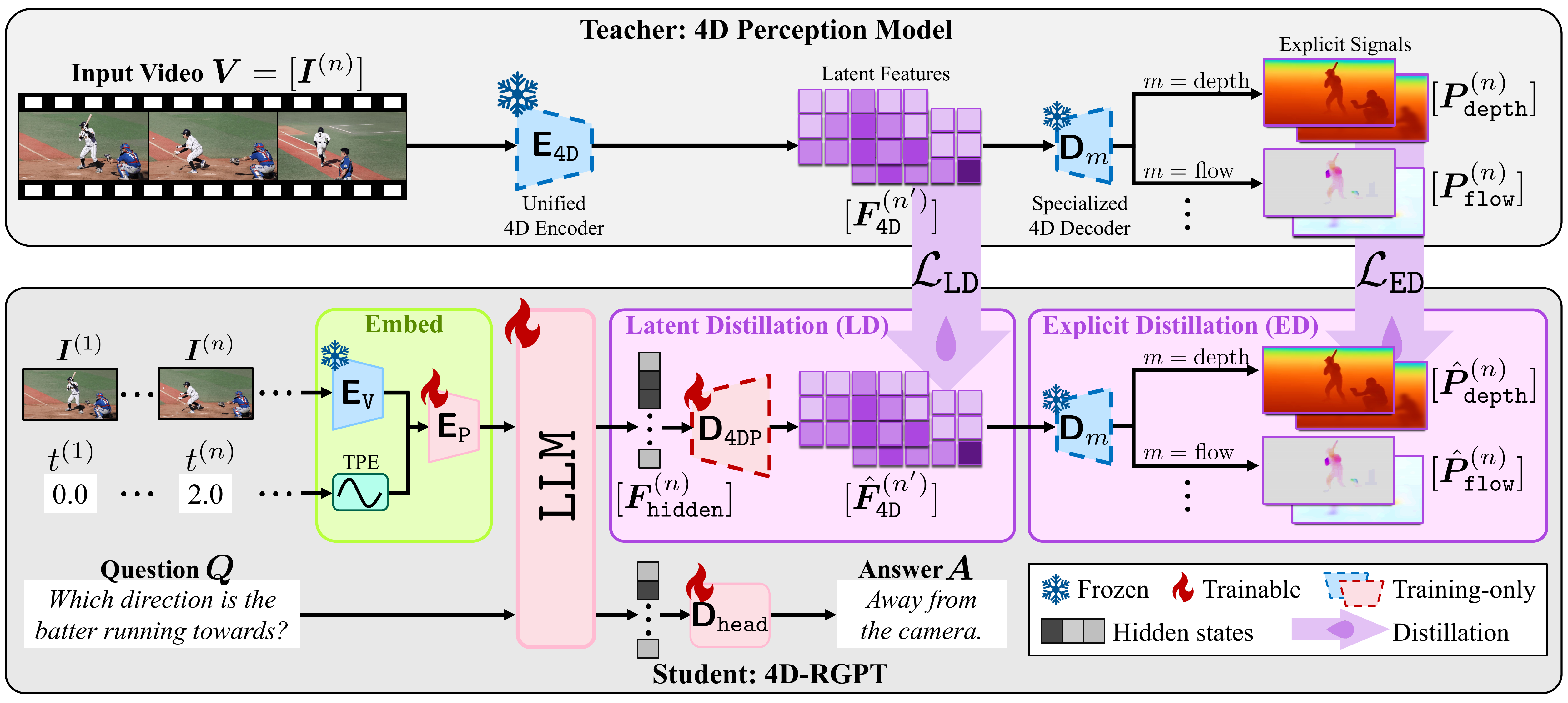}
    \vspace{-0.2cm}
    \caption{
        \textbf{Perceptual 4D Distillation (\ours) framework for \oursabbr.}
        For each frame $\mI^{(i)}$ in $\mV$, \oursabbr~extracts 4D representations through training-only modules, \ie, $\tD_{\tt 4DP}$ and $\tD_m$ for $m \in \gM$.
        This includes both latent features, \ie, $\hat \mF_{\tt 4D}$, and explicit signals, \eg, depth $\hat \mP_{\tt depth}$ or optical flow maps $\hat \mP_{\tt flow}$.
        We also incorporate timestamp positional encodings (TPE) to provide temporal cues for \oursabbr~to be temporally aware.
        In the \ours~framework, the frozen teacher, \ie, 4D perception model, captures 4D expert knowledge from $\mV$.
        It is then distilled to the student \oursabbr~via two strategies.
        (a) \textit{Latent Distillation~(LD)}: We align the latent $\hat \mF_{\tt 4D}$ with the teacher's intermediate 4D embeddings $\mF_{\tt 4D}$.
        (b) \textit{Explicit Distillation~(ED)}: We align the explicit $\hat \mP_{m}$ with the teacher's final 4D signals $\mP_{m}$.
        \oursabbr~is optimized end-to-end using both SFT loss and the distillation losses, \ie, $\gL_{\tt LD}$ and $\gL_{\tt ED}$.
    }
    \label{fig:pipeline}
\end{figure*}

%% file: sections/5_dataset.tex
\input{figures/region4d}

\section{\ourbenchmark}
\label{sec:benchmark}

Recently, there has been significant progress in 3D/4D VQA~\cite{li2025stibench,zhou2025vlm4d,ray2024sat,yang2025mmsi,jia2025omnispatial,fan2025vstibench}.
Several new benchmarks require MLLMs to have depth perception or understand 3D interactions among objects.
However, existing benchmarks do not evaluate MLLMs on 4D region-based understanding in complex, real-world scenarios.
As shown in Tab.~\ref{tab:benchmark_comparison}, they lack the following critical properties:
\begin{itemize}
    \item \textbf{Lack of Dynamic Scenes}:
    Most focus on indoor scenes with minimal object interaction or constrained movement, which do not fully capture the complexity of real-world object manipulation and dynamic changes.
    \item \textbf{Lack of Region Prompting}:
    Region prompts allow controlled and intuitive user queries in VQA.
    Without this ability, an MLLM's interpretability and usability in practical applications are hindered.
\end{itemize}

To address these gaps, we introduce \textit{\textbf{\ourbenchmark}}~(see the rightmost example in Fig.~\ref{fig:region4d}), a novel benchmark that challenges MLLMs with region-level 4D VQA, where depth and temporal perception are critical.

\myparagraph{Task Formulation.} Given an input video $\mV = [\mI^{(n)}]_{n=1:N}$ of $N$ frames, a region-prompted 4D question $\mQ$, and a set of region masks $\mM$ describing the objects of interest in $\mQ$ in $\mI^{(1)}$, the task is to respond with the correct or most suitable answer from a set of options.

\myparagraph{Benchmark.} We curate \ourbenchmark~based on existing non-region-based 4D VQA benchmarks, \ie, STI-Bench~\cite{li2025stibench} and VLM4D~\cite{zhou2025vlm4d}.
Our pipeline (Fig.~\ref{fig:region4d}) employs a hybrid automated and human-verified process to transform conventional VQ pairs into highly specific region-prompted questions.

The process begins with a non-region-prompted 4D VQA.
In the example of Fig.~\ref{fig:region4d}, we are given a video of two persons and a drone with the query question ``How did the person move the drone?''
First, we use Qwen2.5-VL~\cite{alibab2025qwen25vl} to perform keyword extraction (\textbf{Extract}) and identify objects of interest from the query question, \eg, the {\it person} and the {\it drone}.
While videos from some sources, \eg, DAVIS~\cite{ponttuset2017davis}, provide annotations of object masks, other real-world videos lack such detailed annotations.
Hence, we leverage state-of-the-art object detection and segmentation models, \ie, GroundingDINO~\cite{liu2024groundingdino} and SAM2~\cite{ravi2024sam2}, to generate accurate object masks (\textbf{Detect \& Segment}) for the identified objects of interest.
We then apply the segmentation masks with their corresponding keywords onto the video frame to generate an image with \textbf{Set-of-Marks}~\cite{yang2023som}.
This serves as an intermediate and potential portrayal of the region-prompted QA before the final step of checking correctness.

Since the objects of interest can be non-unique (\eg, multiple persons) and segmentation masks can be noisy, ensuring correct association between extracted keywords and found regions is critical.
We check correctness with both automated and human-in-the-loop processes.
We use Qwen2.5-VL~\cite{alibab2025qwen25vl} to automatically match the generated region marks to the entities in the question (\textbf{Matching}).
Finally, human annotators verify and correct any mismatches (\textbf{Verification}). We also trim videos to ensure all RoIs are visible in the first frame.

This concludes our region prompting process. The original VQA is transformed into \ourbenchmark~format, where entities are replaced by region tokens,
\eg, ``How did $\textcolor{darkgreen}{\langle R1\rangle}$ move $\textcolor{red}{\langle R2\rangle}$?'' with their corresponding region masks.

\myparagraph{Statistics.}
Our \ourbenchmark~benchmark consists of 1,517 region-prompted VQAs.
Each question is a multiple-choice problem with four to five answer options.
The benchmark provides region-prompted challenges to semantic and numerical 4D understanding in both static and dynamic scenes.
The static split (418 VQAs) includes 3 categories:
(1) Dimension Measurement; (2) 3D Video Grounding; and (3) Spatial Relation.
The dynamic split (1,098 VQAs) includes 6 categories:
(1) Counting objects; (2) Translational movement; (3) Rotational movement; (4) False Positive detection; (5) Speed \& Acceleration estimation; and (6) Displacement \& Path Length measurement.
We provide more details for each question type in the supplementary material.

%% file: figures/region4d.tex
\begin{figure*}[t]
    \centering
    \includegraphics[width=1.0\textwidth]{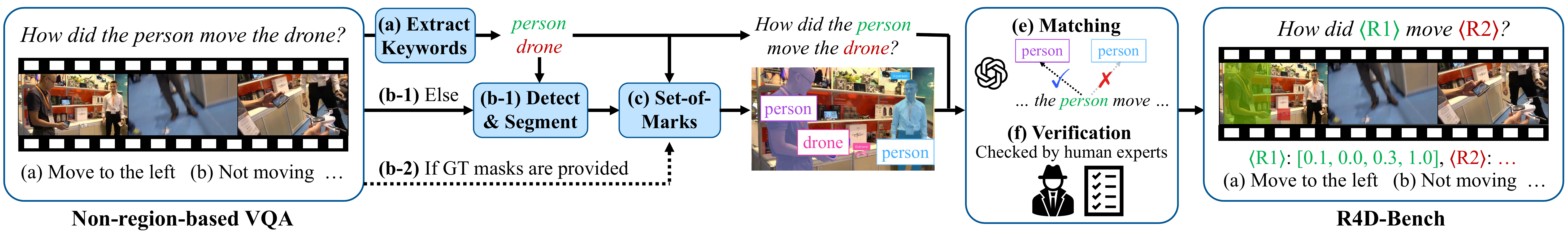}
    \vspace{-0.6cm}
    \caption{
        \textbf{Curation pipeline of our \ourbenchmark.}
        Given existing non-region 4D VQA benchmarks, we (a) first extract the noun keywords from the question as candidates for objects of interest.
        (b) Next, if ground truth segmentation masks are provided, we use them for step (d).
        Otherwise, we use off-the-shelf GroundingDINO~\cite{liu2024groundingdino} and SAM2~\cite{ravi2024sam2} to extract segmentation masks for each object of interest.
        (c) We generate a SoM~\cite{yang2023som} image for the first frame.
        (d) We prompt Qwen-2.5VL~\cite{alibab2025qwen25vl} with the SoM image and the processed question to match the objects referred to in the question with the regions.
        (e) Finally, the generated matching results are verified by human experts.
    }
    \label{fig:region4d}
    \vspace{-0.3cm}
\end{figure*}

%% file: sections/6_exp.tex
\section{Experiments}
\label{sec:exp}

\subsection{Experiment Setup}
\label{sec:exp_setup}

\myparagraph{Benchmarks.}
We evaluate our \oursabbr~on various 4D VQA benchmarks, including our \ourbenchmark~and existing ones, \ie, STI-Bench~\cite{li2025stibench}, VLM4D-real~\cite{zhou2025vlm4d}, OmniSpatial~\cite{jia2025omnispatial}, MMSI-Bench~\cite{yang2025mmsi}, SAT~\cite{ray2024sat}, and VSTI-Bench~\cite{fan2025vstibench}.
Please note that the first four benchmarks are testing-only benchmarks and are disjoint from our training data.
Apart from the numerical questions in VSTI-Bench, where we report relative accuracy,
we report the multiple-choice accuracy for all other benchmarks.

\begin{table}[t!]
    \input{tables/main_comparison}
\end{table}

\begin{table*}[t!]
    \begin{minipage}[t]{0.67\linewidth}
        \input{tables/main_region4d}
    \end{minipage}
    \hfill
    \begin{minipage}[t]{0.30\linewidth}
        \input{tables/ablation_naive}
    \end{minipage}
    \vspace{-0.0cm}
\end{table*}

\myparagraph{Comparison Models.}
We compare our \oursabbr~with various proprietary MLLMs, \eg, GPT-4o~\cite{openai2024gpt4o}, GPT-5~\cite{openai2025gpt5}, Gemini-2.5-Pro~\cite{comanici2025gemini}; open-source generalized MLLMs, \eg, Qwen2.5-VL~\cite{alibab2025qwen25vl}; and recent 3D/4D specialized MLLMs, \eg, SpatialReasoner~\cite{ma2025spatialreasoner}, ViLaSR~\cite{wu2025vilasr}, and SpaceR~\cite{ouyang2025spacer}.

\myparagraph{Architecture.} We select a SOTA open-source generalized MLLM, NVILA-Lite-8B~\cite{liu2025nvila}, as our MLLM backbone, which uses SigLIP~\cite{zhai2023siglip} as the $\tE_{\tt V}$ and Qwen2~\cite{team2024qwen2} as the ${\tt LLM}$.
For the 4D perception model $\tE_{\tt 4D}$ and $\tD_m$, we follow the exact architecture and weights of L4P~\cite{badki2025l4p}.
We document training setups in the supplementary material.

\subsection{Main Results}
We present the effectiveness of \oursabbr~in Tab.~\ref{tab:overall-comparison} and Tab.~\ref{tab:main_region4d}, showing improvements over baseline MLLMs.

\myparagraph{Non-region-based 4D VQA.}
In~\tabref{tab:overall-comparison}, we evaluate \oursabbr~on several non-region-level 3D/4D VQA benchmarks, including input modalities of both images and videos.
We compare with various state-of-the-art proprietary MLLMs, open-source general MLLMs, and recent 3D/4D MLLMs.
\oursabbr~consistently improves over the baseline NVILA-Lite-8B by a large margin across all benchmarks, especially on VLM4D~\cite{zhou2025vlm4d} and VSTI-Bench~\cite{fan2025vstibench}.
Compared to other MLLMs with similar model sizes, \oursabbr~achieves SOTA performance over open-source MLLMs and competitive performance with GPT-4o~\cite{openai2024gpt4o}.
Please note that SpatialReasoner~\cite{ma2025spatialreasoner}, ViLaSR~\cite{wu2025vilasr}, and SpaceR~\cite{ouyang2025spacer} are all trained with RL to further boost accuracy.

\myparagraph{\ourbenchmark.}
In~\tabref{tab:main_region4d}, we present quantitative comparisons of our \oursabbr~on \ourbenchmark~against other MLLMs.
For fair comparison, we use SoM~\cite{yang2023som} to indicate the regions of interest for all MLLMs.
Additionally, for all open-source MLLMs and \oursabbr, we use the same number of sampled frames, \ie, 16 frames.
We observe that although SpaceR~\cite{ouyang2025spacer} outperforms Qwen2.5-VL~\cite{alibab2025qwen25vl} in~\tabref{tab:overall-comparison}, it falls behind on \ourbenchmark,
suggesting that SpaceR is highly tuned for non-region VQA and its region understanding is weakened.
Overall, \oursabbr~achieves the best performance among all open-source MLLMs by at least $1.6\%$ on average and $2.6\%$ on the dynamic split.

In~\figref{fig:main_qual}, we showcase two cases of \oursabbr~against other MLLMs on \ourbenchmark.
In both cases, the regions of interest are constantly moving.
Only \oursabbr~effectively perceives the 4D dynamics and provides the correct answers.
\input{figures/qual_comparison}

\subsection{Ablation Studies}
\label{sec:exp_analysis}

To justify our various designs, we conduct extensive ablation studies and analysis.
For most experiments in this subsection, we report results on STI-Bench~\cite{li2025stibench} and the static and dynamic question subsets of \ourbenchmark.
Without specific notes, we use the same training data, and
all other components are kept identical unless specified.

\myparagraph{Alternative Strategies.}
Besides \ours, there are other strategies to utilize 4D conversation data or the latent feature $\mF_{\tt 4D}$ from the 4D perception models to enhance MLLMs' 4D understanding.
First, denoted as {\it 4D-SFT}, we apply solely SFT to the entire MLLM without access to $\mF_{\tt 4D}$.
Additionally, there are two straightforward ways to leverage $\mF_{\tt 4D}$.
Denoted as {\it 4D-Concat}, we directly concatenate $\mF_{\tt 4D}$ with the 2D visual features $\mE_{\tt V}(\mV)$. We note that this requires additional training on $\tE_{\tt P}$ as the dimension differs from the original visual features.
On the other hand, denoted as {\it 4D-PE}, we project $\mF_{\tt 4D}$ to positional encodings (PE) for the visual features, similar to the spatial PE proposed in SR-3D~\cite{cheng2025sr3d}.

As shown in~\tabref{tab:ablation_naive}, apart from {\it 4D-PE}, both {\it 4D-SFT} and {\it 4D-Concat} improve over the {\it Zero-shot} baseline.
However, they all fall short compared to \ours.
Moreover, {\it 4D-Concat} and {\it 4D-PE} require additional inference costs as they need to compute $\mF_{\tt 4D}$ for each input during inference.
In comparison, \ours~requires solely training-only 4D perception modules, making \oursabbr~as efficient as {\it Zero-shot} during inference.

\begin{table}[t!]
    \input{tables/ablation_distillation}
\end{table}

\myparagraph{Perceptual 4D Distillation.}
To validate the effectiveness of \ours, we experiment with various distillation strategies used
in latent distillation ($\gL_{\tt LD}$ in~\equref{eq:loss_ld}) and explicit distillation ($\gL_{\tt ED}$ in~\equref{eq:loss_ed}).
In~\tabref{tab:ablation_distill}, we ablate different combinations of distillation on $\hat \mF_{\tt 4D}$ and $\hat \mP_{m}$.

We first observe that applying $\gL_{\tt LD}$ alone ({\it LD-only}) improves the performance over the {\it Zero-shot} baseline by $2.3\%$ on \ourbenchmark.
For $\gL_{\tt ED}$, adding more $m \in \gM$ incrementally improves the performance steadily, with $m = {\tt depth}$ and $m = {\tt flow}$ being the most effective ones (see {\it LD+D} and {\it LD+D+F}).
While $\gL_{\tt ED}$ alone ({\it ED-only}) also improves the performance on \ourbenchmark~by $1.9\%$, combining both ({\it LD+ED}) achieves the best average performance, showing the complementary benefits of both LD and ED.

\input{figures/4DD_visual}

\myparagraph{4D Perception Visualization.}
In~\figref{fig:4DD_visual}, we visualize the progress of how \oursabbr~learns to extract 4D signals through \ours.
We show a video from our training set~\cite{lu2025robofac} with extracted $\hat \mP_{\tt depth}$ at various steps.
$\hat \mP_{\tt depth}$ is barely meaningful at first but gradually captures the 3D structure of the scene as training proceeds.
This indicates that \ours~successfully distills 4D perception capabilities into \oursabbr.

\begin{table}[t!]
    \input{tables/ablation_time}
\end{table}

\myparagraph{Timestamp Positional Encoding (TPE).}
MLLMs often struggle with temporal perception when no explicit time cues are provided.
We conduct a controlled toy experiment to validate this observation by curating a simple benchmark with VQAs that require temporal perception, such as ``\textit{How many seconds have passed in the input video?}''
We observe that NVILA-Lite-8B~\cite{liu2025nvila} is naively guessing the answers, resulting in accuracy close to random guessing.
This problem is further exacerbated by the inconsistency among multiple sources of data with different frame rates.
We detail the toy experiment in the supplementary material.

Without introducing additional modules, we test two simple solutions to provide explicit temporal cues to MLLMs.
First, denoted as {\it \ours+mark}, we add explicit time marks similar to SoM~\cite{yang2023som} on each $\mI^{(n)}$, such as burned-in text showing the timestamp, \eg, ``$t^{(n)}$ s''
Second, denoted as {\it \ours+prompt}, we add explicit time information in $\mQ$, such as ``\textit{The following video frames are sampled from a video 19 seconds long and recorded at 30 frames per second.}''

Both {\it \ours+mark} and {\it \ours+prompt}, as shown in Tab.~\ref{tab:ablation_time}, can improve 4D VQA performance.
However, they require additional data preprocessing, distract MLLMs from the main visual and textual content, and do not generalize well to region-level settings, \ie, \ourbenchmark.
Our {\it \ours+TPE} consistently improves performance across both benchmarks, as shown in the last row of Tab.~\ref{tab:ablation_time}.

\begin{table}[t!]
    \input{tables/ablation_model}
\end{table}

\myparagraph{Architecture Design.}
In~\tabref{tab:ablation_model}, we ablate different designs on whether $\tE_{\tt V}$, $\tE_{\tt P}$, or {\tt LLM} is trainable or frozen.
Our {\it Tune-P+LLM} achieves the best performance by tuning both $\tE_{\tt P}$ and {\tt LLM}, while keeping $\tE_{\tt V}$ frozen.
This is likely because $\tE_{\tt P}$ requires finetuning for TPE and \ours~works best on {\tt LLM}.

%% file: tables/main_comparison.tex
\centering
\captionof{table}{
    \textbf{Evaluation on non-region-level 3D / 4D benchmarks.}
    We report the average multiple-choice accuracy $(\uparrow)$ on each benchmark.
    For simplicity, we use the following abbreviations:
    STI (STI-Bench~\cite{li2025stibench}),
    V4D (VLM4D-real~\cite{zhou2025vlm4d}),
    MMSI (MMSI-Bench~\cite{yang2025mmsi}),
    OS (OmniSpatial~\cite{jia2025omnispatial}),
    and
    VSTI (VSTI-Bench~\cite{fan2025vstibench}).
    \label{tab:overall-comparison}
}
\setlength{\tabcolsep}{3pt}
\resizebox{\linewidth}{!}{
\begin{tabular}{l ccc ccc c}
    \specialrule{.15em}{.05em}{.05em}
    {Methods}
    & STI
    & V4D
    & MMSI
    & OS
    & SAT
    & VSTI
    \\
    \midrule
    GPT-4o~\cite{openai2024gpt4o}
    & 34.8 & 60.0 & 30.3 & 47.8 & 57.5 & 38.2
    \\
    GPT-5~\cite{openai2025gpt5}
    & 39.3 & - & 40.7 & 59.9 & - & -
    \\
    Gemini-2.5-Pro~\cite{comanici2025gemini} & 41.4 & 63.5 & 36.9 & 55.4 & - & -
    \\
    Gemini-1.5-Pro~\cite{team2024gemini} & - & - & - & - & 64.8 & -
    \\
    \midrule
    InternVL2.5-8B~\cite{chen2024internvl25} & - & 42.4 & 28.7 & - & - & -
    \\
    Qwen2.5-VL-7B~\cite{alibab2025qwen25vl} & 32.1 & 43.3 & 25.9 & \underline{39.2} & - & -
    \\
    VideoLLaMA3-7B~\cite{zhang2025videollama3} & 35.2 & 46.5 & - & - & - & -
    \\
    LLaVA-Video-7B~\cite{zhang2024llavavideo} & - & - & - & - & 53.5 & -
    \\
    LLaVA-OneVision-7B~\cite{li2024llava}
    & 29.0 & 36.0 & 24.5 & 35.7 & 41.7 & -
    \\
    LLaVA-NeXT-Video-7B~\cite{liu2024llavanext}
    & 29.9 & - & 26.8 & - & - & 40.0
    \\
    \midrule
    VLM-3R-7B~\cite{fan2025vstibench}
    & - & - & - & - & - & \underline{58.8}
    \\
    LLaVA-Video-7B + SAT~\cite{ray2024sat} & - & - & - & - & \underline{63.4} & -
    \\
    ViLaSR-7B~\cite{wu2025vilasr}
    & 33.4 & 46.9 & \underline{30.2} & - & - & -
    \\
    SpatialReasoner-7B~\cite{ma2025spatialreasoner}
    & 31.0 & 43.4 & 22.7 & - & - & -
    \\
    SpaceR-7B~\cite{ouyang2025spacer}
    & \underline{37.0}
    & \underline{51.3} & 28.8 & - & 47.8 & -
    \\
    \midrule
    NVILA-Lite-8B~\cite{liu2025nvila} & 33.8 & 46.5 & 31.3 & 37.2 &  62.0 & 45.2
    \\
    \oursrow
    & \bf 37.6 & \bf 52.7 & \bf 33.3 & \bf 40.4 & \bf 64.7 & \bf 59.1
    \\
    \noalign{\vskip-0.5pt}
    \oursrow
    \multirow{-2}{*}{\oursabbr-8B (Ours)}
    & \improve{+3.8} & \improve{+6.2} & \improve{+2.0} & \improve{+3.2} & \improve{+2.7} & \improve{+13.9}
    \\
    \specialrule{.15em}{.05em}{.05em}
\end{tabular}
}

%% file: tables/main_region4d.tex
\vspace{-0.0cm}
\centering
\captionof{table}{
    \textbf{Evaluation on \ourbenchmark.}
    We report performance on the static split (\tcbox[lightgraybox]{\bf Sta}), the dynamic split (\tcbox[lightpurplebox]{\bf Dyn}), and all 9 tasks of \ourbenchmark.
    For simplicity, we abbreviate them as follows:
    3D \textbf{V}ideo \textbf{G}rounding (\tcbox[lightgraybox]{VG});
    \textbf{D}imension \textbf{M}easurement (\tcbox[lightgraybox]{DM});
    \textbf{S}patial \textbf{R}elationship (\tcbox[lightgraybox]{SR});
    \textbf{R}otational (\tcbox[lightpurplebox]{R});
    \textbf{C}ounting (\tcbox[lightpurplebox]{C});
    \textbf{T}ranslational (\tcbox[lightpurplebox]{T});
    \textbf{F}alse \textbf{P}ositive (\tcbox[lightpurplebox]{FP});
    \textbf{S}peed \& \textbf{A}cceleration (\tcbox[lightpurplebox]{SA});
    and
    \textbf{D}isplacement \& \textbf{P}ath Length (\tcbox[lightpurplebox]{DP}).
    \label{tab:main_region4d}
}
\small
\setlength{\tabcolsep}{2pt}
\resizebox{\linewidth}{!}{
\begin{tabular}{lc<{\hskip 8pt}c c c<{\hskip 8pt} c cc cccccccc}
    \specialrule{.15em}{.05em}{.05em}
    {Methods}
    & {\bf Avg}
    & \cellcolor{lightgray}{\bf Sta}   & \cellcolor{lightpurple}{\bf Dyn} &
    & \cellcolor{lightgray}{VG}   & \cellcolor{lightgray}{DM}     & \cellcolor{lightgray}{SR} &
    & \cellcolor{lightpurple}{R}    & \cellcolor{lightpurple}{C}    & \cellcolor{lightpurple}{T} & \cellcolor{lightpurple}{FP}
    & \cellcolor{lightpurple}{SA}   & \cellcolor{lightpurple}{DP}
    \\
    \midrule
    Random
    & 23.4
    & 20.0 & 24.7 &
    & 20.0 & 20.0 & 20.0 &
    & 25.0 & 25.0 & 25.0 & 25.0
    & 20.0 & 20.0
    \\
    \midrule
    GPT-4o~\cite{openai2024gpt4o} &
    42.8  & 30.3     & 47.5 &     & 30.7 & 26.8 & 43.9 && 49.1 & 35.2 & 51.8 & 54.1 & 27.0 & 10.7
    \\
    \midrule
    Qwen2.5-VL-7B~\cite{alibab2025qwen25vl}
    & \underline{40.6}  & \bf 34.1     & \underline{43.1} &
    & 39.1 & 25.7 & 48.8 &
    & 50.0 & 38.4 & 46.6 & 28.9
    & 45.9 & 28.6
    \\
    LLaVA-Video-7B~\cite{zhang2024llavavideo}
    & 39.7  & 26.9     & 44.6      && 23.4 & 28.4 & 36.6 && 46.2 & 30.2 & 50.4 & 33.6 & 48.6 & 35.7
    \\
    \midrule
    ViLaSR-7B~\cite{wu2025vilasr}
    &   39.6  & 31.5     & 42.6      && 34.4 & 24.6 & 48.8 && 46.2 & 42.8 & 51.3 & 3.7  & 43.2 & 17.9
    \\
    SpatialReasoner-7B~\cite{ma2025spatialreasoner}
    &   38.3  & 31.2     & 41.0      && 35.4 & 25.7 & 36.6 && 43.4 & 37.1 & 49.3 & 11.9 & 32.4 & 17.9
    \\
    SpaceR-7B~\cite{ouyang2025spacer}
    & 37.0  & 26.2     & 41.1  &     & 30.7 & 18.0 & 41.5 && 47.2 & 40.3 & 43.8 & 25.9 & 51.4 & 21.4
    \\
    \midrule
    NVILA-Lite-8B~\cite{liu2025nvila}
    & 37.9  & 29.1     & 41.3 &
    & 33.9 & 20.2 & 46.3 &
    & 41.5 & 39.6 & 41.9 & 40.7
    & 45.9 & 32.1
    \\
    \oursrow
    & \bf 42.2
    & \underline{32.9} & \bf 45.7 &
    & 35.1 & 26.3 & 52.2 &
    & 43.1 & 40.1 & 48.7 & 40.2
    & 50.9 & 38.9
    \\
    \noalign{\vskip-0.5pt}
    \oursrow
    \multirow{-2}{*}{\oursabbr-8B (Ours)}
    & \improve{+4.3} & \improve{+3.8} & \improve{+4.4} &
    & \improve{+1.2} & \improve{+6.1} & \improve{+5.9} &
    & \improve{+1.6} & \improve{+0.5} & \improve{+6.8} & -0.5
    & \improve{+5.0} & \improve{+6.8}
    \\
    \specialrule{.15em}{.05em}{.05em}
\end{tabular}
    }

%% file: tables/ablation_naive.tex
\vspace{-0.0cm}
\centering
\captionof{table}{
    \textbf{Alternative strategies for 4D VQA.}
    We compare \ours~with direct SFT ({\it 4D-SFT}) and straightforward designs of incorporating $\mF_{\tt 4D}$ from the 4D perception model, \ie, {\it 4D-Concat} and {\it 4D-PE}.
    For simplicity, we use the same abbreviations as in~\tabref{tab:main_region4d} and STI for STI-Bench~\cite{li2025stibench}.
    \label{tab:ablation_naive}
}
\resizebox{\linewidth}{!}{
\begin{tabular}{l ccc ccc cc}
    \specialrule{.15em}{.05em}{.05em}
    \multirow{2}{*}{Methods} &
    \multirow{2}{*}{$\mF_{\tt 4D}$} &
    \multirow{2}{*}{STI} &
    \multicolumn{3}{c}{\ourbenchmark}
    \\
    \cmidrule(lr){4-6}
    & &
    & Avg & Sta & Dyn
    \\
    \midrule
    {\it Zero-shot}
    & \ccross
    & 33.8
    & 37.9 & 29.1 & 41.3
    \\
    \midrule
    {\it 4D-SFT}
    & \ccross
    & 34.7
    & 40.1 & \underline{32.2} & \underline{43.8}
    \\
    {\it 4D-Concat}
    & \ccheck
    & \underline{34.8}
    & \underline{39.5} & 30.6 & {42.9}
    \\
    {\it 4D-PE}
    & \ccheck
    & 31.3
    & 36.0 & 26.6 & 39.5
    \\
    \midrule
    \oursrow Ours ({\it \ours})
    & \ccheck
    & \bf 37.6
    & \bf 42.2 & \bf 32.9 & \bf 45.7
    \\
    \specialrule{.15em}{.05em}{.05em}
\end{tabular}
}

%% file: figures/qual_comparison.tex
\begin{figure}[t]
    \centering
    \includegraphics[width=\linewidth]{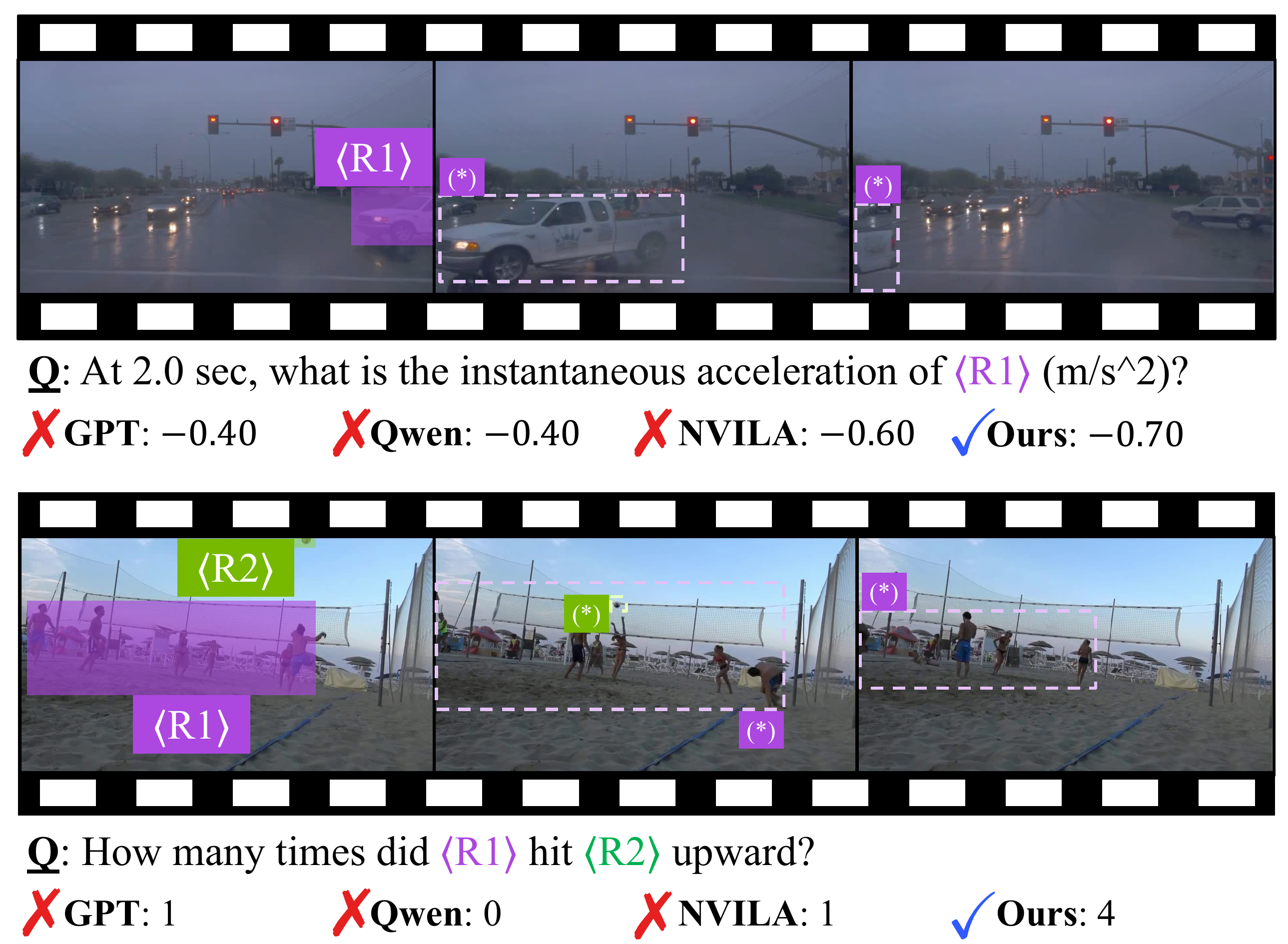}
    \vspace{-0.5cm}
    \caption{
        \textbf{VQA comparison among baseline MLLMs and \oursabbr~on \ourbenchmark.}
        For the baseline MLLMs, we use GPT-4o-20241120~\cite{openai2024gpt4o}, Qwen-2.5VL-7B-Instruct~\cite{alibab2025qwen25vl}, and NVILA-Lite-8B~\cite{liu2025nvila}.
        We note that the regions labeled with \tcbox[tightpurple]{(*)} or \tcbox[tightgreen]{(*)} are not provided in \ourbenchmark; they are visualized for readability.
    }
    \label{fig:main_qual}
    \vspace{-0.5cm}
\end{figure}

%% file: tables/ablation_distillation.tex
\vspace{-0.0cm}
\centering
\captionof{table}{
    \textbf{Analysis of 4D modalities in \ours.}
    We ablate the effectiveness of different combinations of distillation in latent distillation (LD) on $\hat \mF_{\tt 4D}$ and explicit distillation (ED) on $\hat \mP_{m}$.
    For simplicity, we use the same abbreviations as Tab.~\ref{tab:ablation_naive} and \textbf{D}epth (D), \textbf{F}low (F), \textbf{M}otion (M), and \textbf{C}amray (C) for each $m \in \gM$.
    \label{tab:ablation_distill}
}
\resizebox{0.98\linewidth}{!}{
\begin{tabular}{l ccc ccc ccc}
    \specialrule{.15em}{.05em}{.05em}
    \multirow{2}{*}{Methods} &
    \multirow{2}{*}{$\hat \mF_{\tt 4D}$} &
    \multicolumn{4}{c}{$\hat \mP_m$} &
    \multirow{2}{*}{STI} &
    \multicolumn{3}{c}{\ourbenchmark}
    \\
    \cmidrule(lr){3-6}
    \cmidrule(lr){8-10}
    &
    & D & F & M & C
    &
    & Avg & Sta & Dyn
    \\
    \midrule
    {\it Zero-shot}
    & \ccross & \ccross & \ccross & \ccross & \ccross
    & 33.8
    & 37.9 & 29.1 & 41.3
    \\
    \midrule
    {\it LD-Only}
    & \ccheck & \ccross & \ccross & \ccross  & \ccross
    & 34.2
    & 40.2 & 32.0 & 43.3
    \\
    {\it LD+D}
    & \ccheck & \ccheck & \ccross & \ccross  & \ccross
    & 33.4
    & 40.8 & 32.5 & 44.0
    \\
    {\it LD+D+F}
    & \ccheck & \ccheck & \ccheck & \ccross  & \ccross
    & 36.2
    & 41.9 & \bf 33.1 & 45.3
    \\
    {\it LD+D+F+M}
    & \ccheck & \ccheck & \ccheck & \ccheck  & \ccross
    & \underline{36.5}
    & \underline{42.0} & \bf 33.1 & \underline{45.4}
    \\
    {\it ED-Only}
    & \ccross & \ccheck & \ccheck & \ccheck  & \ccheck
    & 35.4
    & 39.8 & 31.5 & 42.9
    \\
    \midrule
    \oursrow Ours {\it (LD+ED)}
    & \ccheck & \ccheck & \ccheck & \ccheck  & \ccheck
    & \bf 37.6
    & \bf 42.2 & \underline{32.9} & \bf 45.7
    \\
    \specialrule{.15em}{.05em}{.05em}
\end{tabular}
}
\vspace{-0.3cm}

%% file: figures/4DD_visual.tex
\begin{figure}[t]
    \centering
    \includegraphics[width=0.9\linewidth]{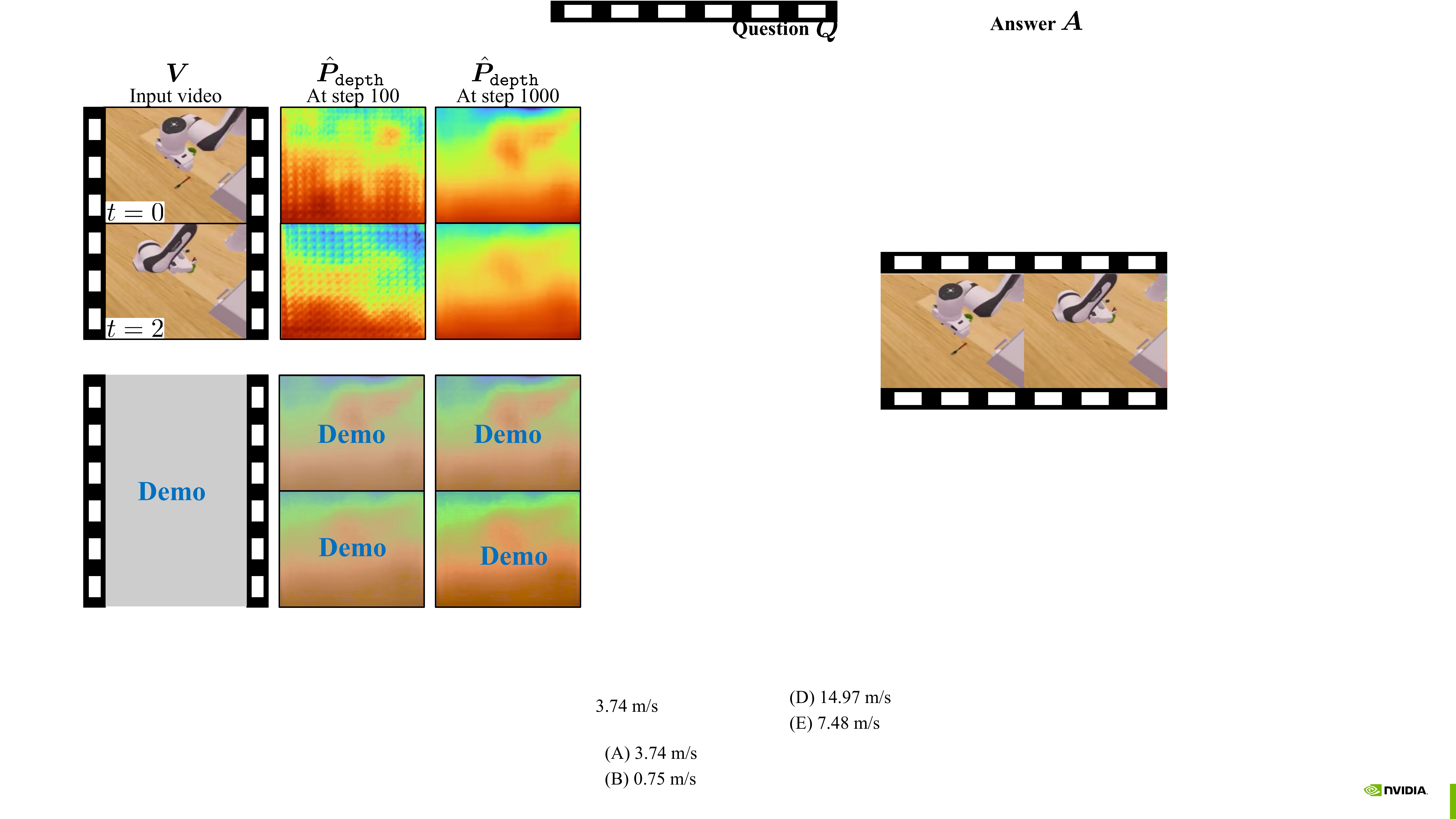}
    \vspace{-0.15cm}
    \caption{
        \textbf{
            Predicted depth maps at different training steps.
        }
        We visualize the progress of $\hat \mP_{\tt depth}$ throughout training.
    }
    \label{fig:4DD_visual}
\end{figure}

%% file: tables/ablation_time.tex
\vspace{-0.0cm}
\centering
\captionof{table}{
    \textbf{Ablation studies on explicit temporal cues.}
    We experiment without and with different choices of explicit time cues.
    For simplicity, we use the same abbreviations as Tab.~\ref{tab:ablation_naive}.
    \label{tab:ablation_time}
}
\resizebox{0.85\linewidth}{!}{
\begin{tabular}{l cccccc ccc c}
    \specialrule{.15em}{.05em}{.05em}
    \multirow{2}{*}{Methods} &
    \multirow{2}{*}{Time cues} &
    \multirow{2}{*}{STI} &
    \multicolumn{3}{c}{\ourbenchmark}
    \\
    \cmidrule(lr){4-6}
    &
    &
    & Avg & Sta & Dyn
    \\
    \midrule
    {\it Zero-shot}
    & \ccross
    & 33.8
    & 37.9 & 29.1 & 41.3
    \\
    \midrule
    {\it \ours} & \ccross
    & 34.8
    & 41.0 & 31.8 & 44.5
    \\
    {\it \ours+mark}
    & marks
    & 35.1
    & 41.1 & 31.5 & 44.7
    \\
    {\it \ours+prompt}
    & prompts
    & \underline{36.1}
    & \underline{41.5} & \underline{32.1} & \underline{45.0}
    \\
    \midrule
    \oursrow Ours ({\it \ours+TPE})
    & TPE
    & \bf 37.6
    & \bf 42.2 & \bf 32.9 & \bf 45.7
    \\
    \specialrule{.15em}{.05em}{.05em}
\end{tabular}
}
\vspace{-0.1cm}

%% file: tables/ablation_model.tex
\vspace{-0.0cm}
\centering
\captionof{table}{
    \textbf{Ablation studies on different training designs in \oursabbr.}
    We ablate different training designs on whether each module is trainable and whether to use LoRA~\cite{hu2022lora}.
    For simplicity, we use the same abbreviations as Tab.~\ref{tab:ablation_naive}.
    \label{tab:ablation_model}
}
\vspace{-0.1cm}
\resizebox{0.99\linewidth}{!}{
\begin{tabular}{l ccc ccc c}
    \specialrule{.15em}{.05em}{.05em}
    \multirow{2}{*}{Methods} &
    \multicolumn{3}{c}{Trainable} &
    \multirow{2}{*}{STI} &
    \multicolumn{3}{c}{\ourbenchmark}
    \\
    \cmidrule(lr){2-4}
    \cmidrule(lr){6-8}
    & $\tE_{\tt V}$ & $\tE_{\tt P}$ & {\tt LLM}
    &
    & Avg & Sta & Dyn
    \\
    \midrule
    {\it Zero-shot}
    & \ccross & \ccross & \ccross
    & 33.8
    & 37.9 & 29.1 & 41.3
    \\
    \midrule
    {\it Tune-All}
    & \ccheck & \ccheck & \ccheck
    & 34.7
    & 38.8 & 30.1 & 42.1
    \\
    {\it Tune-V}
    & \ccheck & \ccross & \ccross
    & 32.3
    & 35.8 & 27.3 & 39.0
    \\
    {\it Tune-P}
    & \ccross & \ccheck & \ccross
    & 34.3
    & 38.6 & 29.8 & 42.0
    \\
    {\it Tune-LLM}
    & \ccross & \ccross & \ccheck
    & 35.4
    & 40.5 & 32.2 & 43.7
    \\
    {\it Tune-LLM-LoRA}
    & \ccross & \ccross & LoRA
    & \underline{37.0}
    & 41.1 & \bf {33.0} & 44.2
    \\
    {\it Tune-P+LLM-LoRA}
    & \ccross & \ccheck & LoRA
    & 36.5
    & \underline{41.4} & 32.8 & \underline{44.7}
    \\
    \midrule
    \oursrow
    Ours ({\it Tune-P+LLM})
    & \ccross & \ccheck & \ccheck
    & \bf 37.6
    & \bf 42.2 & \underline{32.9} & \bf 45.7
    \\
    \specialrule{.15em}{.05em}{.05em}
\end{tabular}
}
\vspace{-0.15cm}

%% file: sections/7_conc.tex
\section{Conclusion}
We show that existing MLLMs struggle with region-level 4D VQA due to
not fully perceiving 4D information.
Without incurring additional inference cost, our \oursabbr~effectively improves MLLMs' 4D perception by learning from a 4D perception model via a novel distillation framework, \ours. Additionally, we introduce a proper benchmark, \ourbenchmark, for this domain, contributing to region-level 4D VQA.
Extensive experiments confirm the effectiveness of our approach on both non-region-level and region-level 4D VQA. 

%% file: sections/8_ack.tex
\section{Acknowledgment}

We would like to express our gratitude to Abhishek Badki, Hang Su, Boyi Li, Ran Tian, Boris Ivanovic, and Marco Pavone for the model and data sharing and fruitful discussions during the 4D-RGPT development. We also appreciate the helpful discussions on problem formulation and potential applications with the VILA team (Hanrong Ye, Hongxu Yin, Yao Lu) and the Metropolis group (Vidya Murali, Varun Praveen, Tomasz Kornuta, Xiaolong Li, Zaid Pervaiz Bhat, Ryan Ji, Adityan Jothi, Thomas Tang, Paris Zhang, Yilin Zhao, Ratnesh Kumar, and Bhanu Pisupati) at NVIDIA.

%% file: sections/X_supp.tex
\clearpage

\section*{Appendix}

\setcounter{section}{0}
\renewcommand{\theHsection}{A\arabic{section}}
\renewcommand{\thesection}{A\arabic{section}}
\renewcommand{\thetable}{A\arabic{table}}
\setcounter{table}{0}
\setcounter{figure}{0}
\renewcommand{\thetable}{A\arabic{table}}
\renewcommand\thefigure{A\arabic{figure}}
\renewcommand{\theHtable}{A.Tab.\arabic{table}}
\renewcommand{\theHfigure}{A.Abb.\arabic{figure}}
\renewcommand\theequation{A\arabic{equation}}
\renewcommand{\theHequation}{A.Abb.\arabic{equation}}

{\bf \noindent The appendix is organized as follows:}
\begin{itemize}
\item In~\secref{supp:details}, we provide implementation and training details for \ours~and \oursabbr, including model architecture, training data, computational resources, and loss functions.
\item In~\secref{supp:benchmark}, we provide the detailed design of \ourbenchmark, including the nine question categories and dataset curation process.
\item In~\secref{supp:results}, we provide additional experimental results, including results with other NVILA variants, analysis of temporal perception capabilities, training data mixture, more qualitative results, and visualizations.
\end{itemize}

\section{Additional Details}
\label{supp:details}

\subsection{Model Architecture}

\myparagraph{MLLM.}
As mentioned in~\secref{sec:exp_setup}, we use NVILA-Lite-8B~\cite{liu2025nvila} as our base MLLM in the main experiments.
NVILA is a unified open-sourced MLLM family that tackles both image and video understanding.

Considering the tradeoff between performance and inference efficiency, there are two groups of NVILA variants, \eg, NVILA (Base) and NVILA-Lite, where the latter is more efficient.
For example, NVILA-Lite uses a $3 \times 3$ downsampling kernel in $\tE_{\tt P}$ while NVILA (Base) uses $2 \times 2$.
We select NVILA-Lite as our base MLLM due to its competitive performance and higher efficiency.

For all NVILA variants, we use their open-sourced weights from {\tt HuggingFace}~\cite{wolf2019HuggingFace}.
Specifically, we use the following checkpoints:
\begin{itemize}
    \item \inlinecode{Efficient-Large-Model/NVILA-Lite-8B};
    \item \inlinecode{Efficient-Large-Model/NVILA-Lite-15B};
\end{itemize}

For the vision encoder (tower) $\tE_{\tt V}$, they use SigLIP~\cite{zhai2023siglip}, specifically \inlinecode{siglip-so400m-patch14-384}.
For the multi-modal projector $\tE_{\tt P}$, they use a 2-layer MLP with a hidden dimension of 4,608.

\myparagraph{4D Perception Model.}
As mentioned in~\secref{sec:exp_setup}, we use L4P~\cite{liu2025nvila} as our 4D perception model.
A 40-layer ViT-based video encoder from VideoMAEv2~\cite{wang2023videomaev2} is adopted for $\tE_{\tt 4D}$, and
DPT~\cite{ranftl2021dpt} is adopted for each $\tD_m$ where $m \in \gM$.
Each $\tD_m$ has the same architecture but different output channels depending on the target modality.
As mentioned in~\secref{sec:background}, the output channels are $1, 2, 1, 6$ for the ${{\tt depth}, {\tt flow}, {\tt motion}, {\tt camray}}$, respectively.
L4P has 1,337M parameters and takes approximately 300ms to process a 16-frame video on an A100 GPU.
Since L4P is only required during training, its 4D signals can be pre-computed and stored offline, adding no inference overhead to \oursabbr.

\myparagraph{\oursabbr.} In \oursabbr, we design a lightweight 4D perception decoder $\tD_{\tt 4DP}$ to efficiently extract 4D perceptual latent from {\tt LLM}'s hidden states.
It is a 3-layer MLP with a hidden dimension of 2,560.
We use GELU~\cite{hendrycks2016gelu} as the activation function between each layer.
For initialization, we use Xavier initialization~\cite{glorot2010xavierinitialization} for all weights and zeros for all biases.
Additionally, \oursabbr~employs Temporal Positional Encoding (TPE) to enhance the temporal understanding of the model.
For TPE (\equref{eq:tpe}), we use $T = 10,000$.

\subsection{Data Mixture}
We provide more details about the training data mixture used in our training.

\myparagraph{VSTI-Bench~\cite{fan2025vstibench}} is a new dataset built upon VSI-Bench~\cite{yang2025vsibench}. While VSI-Bench focuses on the spatial understanding of static 3D scenes,
VSTI-Bench further investigates the spatial-temporal understanding of how spatial relations evolve over time.
We use only the training set of VSTI-Bench and do not use the VSI-Bench.
The videos are sourced from ScanNet~\cite{dai2017scannet} and ScanNet++~\cite{yeshwanth2023scannetplus}.
The training set contains roughly 1.2k unique videos and 130k QA pairs.
A training sample is shown in Fig.~\ref{fig:vstibench_example}.

\input{figures/supp/vstibench_example}

\myparagraph{Wolf~\cite{li2024wolf}} is a large-scale video captioning dataset with high-quality captions generated by VLMs.
Wolf provides detailed captions across three domains: autonomous driving, general scenes, and robotics.
We use the NuScenes~\cite{caesar2020nuscenes} portion of Wolf, \ie, the autonomous driving domain.
We use Llama-3.1-70B-Instruct~\cite{dubey2024llama3} with the template-based text prompts to generate question-answer pairs based on these captions, creating conversational data suitable for 4D VQA training.
The training set contains roughly 5k unique videos and 15k QA pairs.
A training sample is shown in~\figref{fig:wolf_example}.

\input{figures/supp/wolf_example}

\myparagraph{RoboFAC~\cite{lu2025robofac}} is a large-scale dataset for semantic understanding of robotic arm videos, including a training split with simulated robotic arm videos involving various actions.
We adopt it into our training data mixture due to its stable camera views with limited background variations but rich robotic arm movements. It contains roughly 10k unique videos and 65k conversations. A training sample is shown in~\figref{fig:robofac_example}.

\input{figures/supp/robofac_example}

\myparagraph{SAT~\cite{ray2024sat}} is an image-based VQA dataset.
Though it is image-based, we consider it helpful for 4D VQA training due to its relevance on dynamic scene understanding across images.
The training set contains roughly 190k unique simulated images and 170k QA pairs.
A training sample is shown in Fig.~\ref{fig:sat_example}.

\input{figures/supp/sat_example}

\subsection{Training Details}
Our training starts from the pre-trained NVILA weights with an initial learning rate of $1\mathrm{e}{-5}$.
We use a cosine learning rate scheduler with a warmup ratio of 0.03. We train on a multi-node cluster comprising 8 nodes.
Each node has NVIDIA A100-SXM4-80GB GPUs and an AMD EPYC 7J13 64-Core Processor CPU.
The total batch size is 1,024. We train for 5 epochs over approximately 12 hours.

\myparagraph{Losses.}
As mentioned in~\secref{sec:distillation}, we train our model with both SFT loss $\gL_{{\tt SFT}}$ and \ours~loss,
\ie, latent distillation loss $\gL_{{\tt LD}}$ and explicit distillation loss $\gL_{{\tt ED}}$.
Specifically, our total loss is
\begin{equation}
    \gL = \gL_{\tt SFT} + \alpha \gL_{\tt LD} + \beta \gL_{\tt ED},
\label{eq:total_loss}
\end{equation}
where $\alpha$ and $\beta$ are hyperparameters to balance the three loss terms.
We set $\alpha = 0.5$ and $\beta = 0.1$.

In Eq.~\ref{eq:loss_ld}, we set $\Delta_{\tt LD}$ to be the Smooth-L1 distance function.
In Eq.~\ref{eq:loss_ed}, we set each $\Delta_{m}$ to be the Smooth-L1 distance function and $\lambda_m$ to be $1.0, 0.1, 0.05, 0.05$ for $m \in \{ {\tt depth}, {\tt flow}, {\tt motion}, {\tt camray} \}$, respectively.

\section{\ourbenchmark}
\label{supp:benchmark}
We provide more details about \ourbenchmark, including the 9 question categories (\secref{supp:categories}) and dataset curation process (\secref{supp:curation}).

\input{sections/supp/r4d_process}
\input{sections/supp/r4d_category}

\section{Additional Results}
\label{supp:results}

\myparagraph{More NVILA variants.}
In Tab.~\ref{tab:more_nvila_nonregion} and Tab.~\ref{tab:more_nvila_region4d}, we provide additional results using NVILA-Lite-15B as the base MLLM on non-region-based 4D VQA and \ourbenchmark, respectively.
We observe consistent performance improvements across various benchmarks.

\begin{table}[ht]
    \input{tables/supp/more_nvila_nonregion}

\end{table}

\begin{table}[ht]
    \input{tables/supp/more_nvila_region4d}
\end{table}

\myparagraph{Temporal Perception.}
As discussed in~\secref{sec:4d-rgpt} and \secref{sec:exp_analysis}, we observe that MLLMs tend to struggle with temporal perception.
To demonstrate such a deficiency, we conduct a toy experiment.
As shown in Fig.~\ref{fig:timebench}, we curate \textit{TimeBench}, a simple set of VQA questions that require temporal perception of input frames, such as ``\textit{How many seconds have passed in the input video?}''.
All videos are acquired from the STI-Bench~\cite{li2025stibench} and VLM4D~\cite{zhou2025vlm4d}.
We note that these two benchmarks have 4 different frame rates, ranging from 10 to 30, as shown in Tab.~\ref{tab:benchmark_comparison}.
This makes it even more challenging for MLLMs to infer time duration.
To avoid ambiguity in answers, we provide 4 extra options for each question, ranging from $0.25\times$ to $4\times$ of the actual time duration.

\begin{figure}[ht]
    \input{figures/timebench}
    \label{fig:timebench}
\end{figure}

\begin{table}[ht]
    \input{tables/supp/timebench}
\end{table}

{\it Zero-shot} and {\it \ours} in Tab.~\ref{tab:supp_timebench} show that without cues, MLLMs struggle to know how much time has passed in the input frames.
The baselines are naively guessing the answers, resulting in an accuracy close to random guessing (20\%).
This problem is further exaggerated by the inconsistency that different sources of training data and evaluation benchmarks have different frame rates.

We observe that both {\it \ours+mark} and {\it \ours+prompt} can greatly improve the performance on \textit{TimeBench}, which is expected since they provide explicit temporal cues.
However, they require additional data preprocessing and distract MLLMs from the main visual and textual content.
This toy experiment inspires us to develop methods that can provide temporal cues without modifying the input data, \ie, our TPE.

\myparagraph{Training Data Mixture.}
We incrementally add different datasets to analyze their contributions.
In Tab.~\ref{tab:increment}, we observe that compared to the {\it Zero-shot} baseline, adding the training data gradually improves the performance on both non-region-based (STI-Bench) and region-based 4D VQA (\ourbenchmark).
Though SAT~\cite{ray2024sat} is an image-based VQA dataset, adding it also brings moderate performance gains.

\begin{table}[ht]
    \input{tables/supp/dataset_increment}
\end{table}

\clearpage

\myparagraph{More Qualitative Results.}
Following the format in Fig.~\ref{fig:main_qual}, we provide additional qualitative results on \ourbenchmark~in Fig.~\ref{fig:more_qual_4}, Fig.~\ref{fig:more_qual_1}, Fig.~\ref{fig:more_qual_2}, and Fig.~\ref{fig:more_qual_3}.
\input{figures/supp/more_qual}

\myparagraph{More $\hat \mP_m$ Visualizations.}
In Fig.~\ref{fig:more_depth_vis}, we provide additional visualizations of the \oursabbr~explicit signals $\hat \mP_{m}$ at different training steps.
In earlier steps, we observe inaccurate predictions with grid-like structures.
We hypothesize that this is due to the tokenization process in hidden states of the {\tt LLM} transformer, \ie, $\mF_{\tt hidden}$.
However, as training proceeds, the grid-like structures gradually diminish, leading to smoother and more reasonable predictions.
We demonstrate that our \oursabbr~can effectively learn to extract explicit 4D perceptual signals through the training of \ours.

\input{figures/supp/more_depth_vis}

\myparagraph{Limitations.}
\oursabbr~can still produce suboptimal results, particularly in questions requiring precise numerical estimation, \eg, exact speed or displacement values, as illustrated in several failure cases in Fig.~\ref{fig:more_qual_1}--\ref{fig:more_qual_4}.
We attribute this to the lack of step-by-step reasoning during training.

%% file: figures/supp/vstibench_example.tex
\begin{figure}[t]
    \centering
    \includegraphics[width=\linewidth]{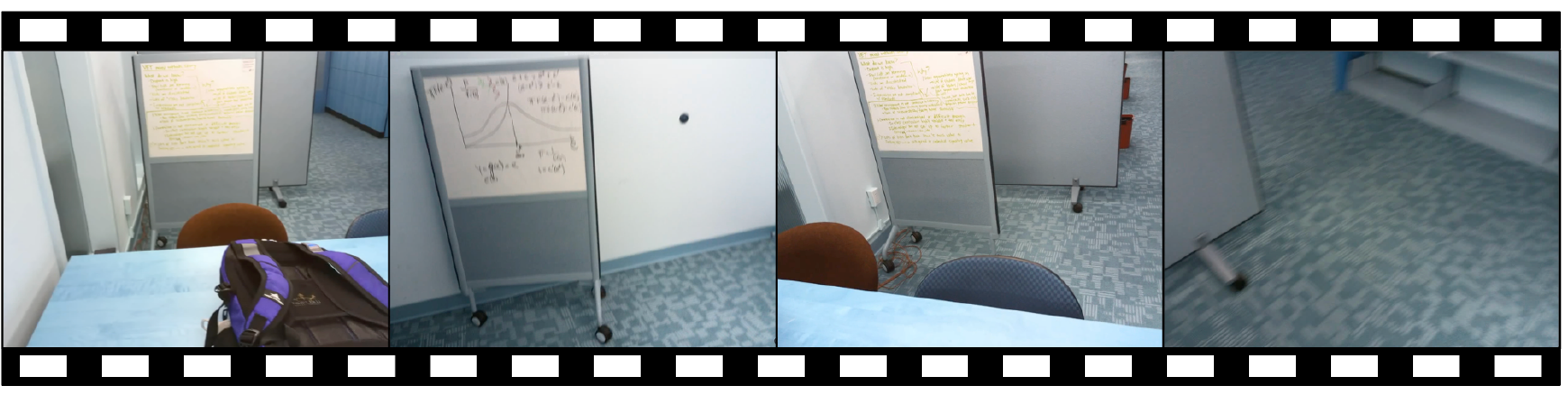}
    \caption{
        \textbf{An example from VSTI-Bench~\cite{fan2025vstibench} training data}.
        The corresponding conversation is as follows:
        (1) \textit{User}: ``These are frames of a video. Approximately how far (in meters) did the camera move between frame 14 and frame 20 of 32? Please answer the question using a single word or phrase.'';
        (2) \textit{GPT}: ``1.6''.
    }
    \vspace{-0.3cm}
    \label{fig:vstibench_example}
\end{figure}

%% file: figures/supp/wolf_example.tex
\begin{figure}[ht]
    \centering
    \includegraphics[width=\linewidth]{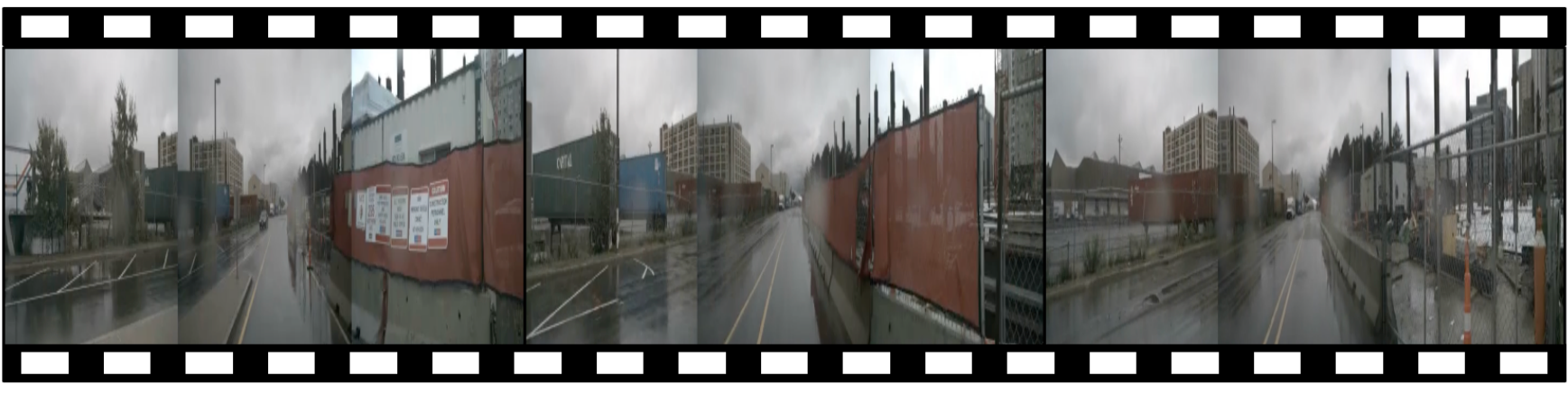}
            \vspace{-0.6cm}
    \caption{
        \textbf{An example from Wolf~\cite{li2024wolf} training data}.
        The corresponding conversation is as follows:
        (1) \textit{User}: ``What traffic participants are around the ego-centric vehicle in the video? Please count and list all of them.'';
        (2) \textit{GPT}: ``1 car is near the ego vehicle's expected path. 1 traffic cone is in the ego vehicle's lane and potentially blocking the ego vehicle. 4 barriers are in the ego vehicle's lane and potentially blocking the ego vehicle.''.
    }
        \vspace{-0.2cm}
    \label{fig:wolf_example}
\end{figure}

%% file: figures/supp/robofac_example.tex
\begin{figure}[t]
    \centering
    \includegraphics[width=\linewidth]{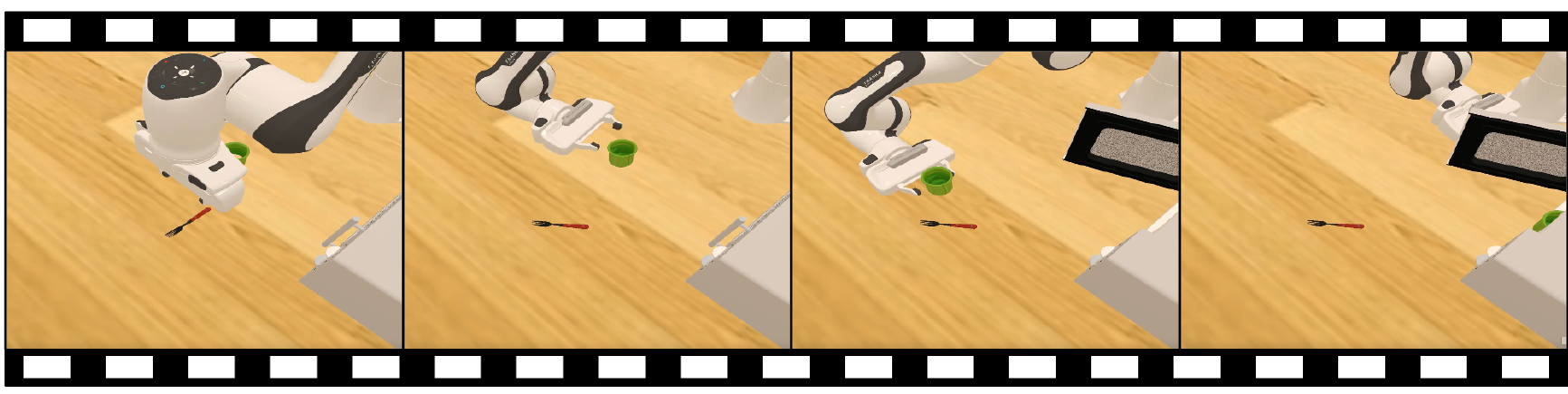}
    \vspace{-0.6cm}
    \caption{
        \textbf{An example from RoboFAC~\cite{lu2025robofac} training data}.
        The corresponding conversation is as follows:
        (1) \textit{User}: ``From the video, what task is the robot engaged in?'';
        (2) \textit{GPT}: ``Put the fork in the cup and put them in the microwave.''.
    }
        \vspace{-0.5cm}
    \label{fig:robofac_example}
\end{figure}

%% file: figures/supp/sat_example.tex
\begin{figure}[t]
    \centering
    \begin{minipage}{0.48\linewidth}
        \centering
        \includegraphics[width=\linewidth,height=0.8\linewidth,keepaspectratio=false]{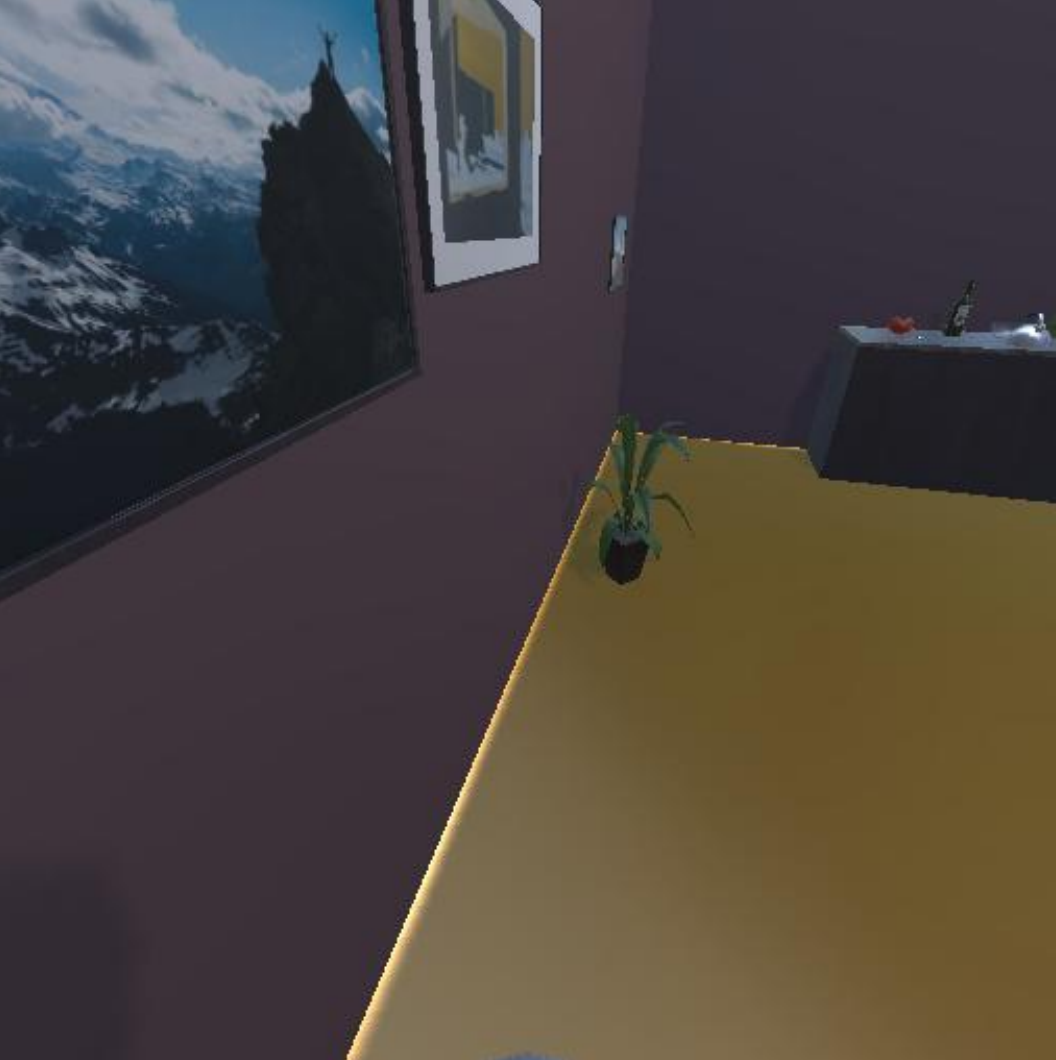}
        \subcaption{First frame.}
    \end{minipage}
    \hfill
    \begin{minipage}{0.48\linewidth}
        \centering
        \includegraphics[width=\linewidth,height=0.8\linewidth,keepaspectratio=false]{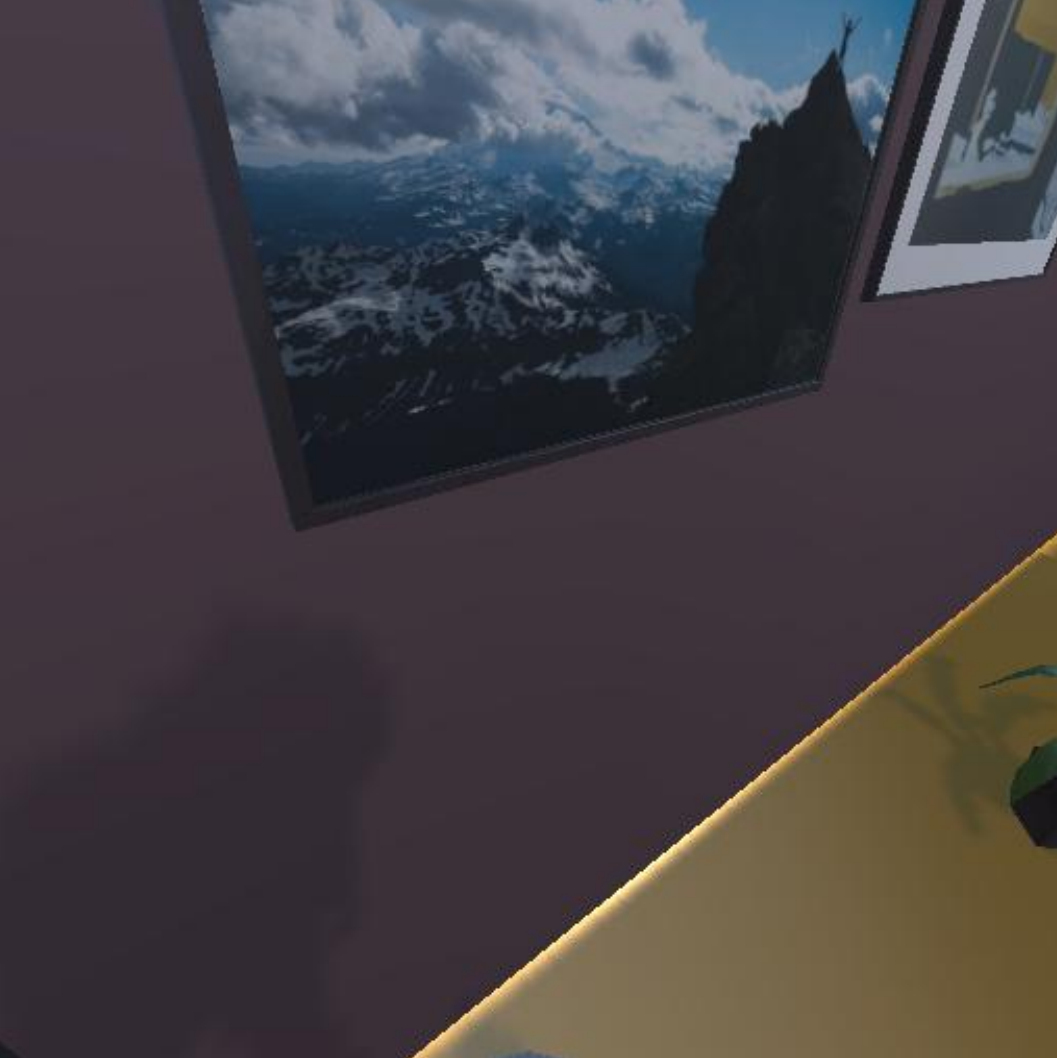}
        \subcaption{Second frame.}
    \end{minipage}
         \vspace{-0.25cm}
    \caption{
        \textbf{An example from SAT~\cite{ray2024sat} training data.}
        The corresponding conversation is as follows:
        (1) \textit{User}:
        ``Were any of the objects in the initial frame that you can still see in the second frame moved from their original positions?
        Options: ['green tapered square potted houseplant was moved right and towards the camera in the first frame', 'green tapered square potted houseplant was moved left and away from the camera in the first frame']'';
        (2) \textit{GPT}: ``green tapered square potted houseplant was moved right and towards the camera in the first frame.''.
    }
        \vspace{-0.5cm}
    \label{fig:sat_example}
\end{figure}

%% file: sections/supp/r4d_process.tex
\subsection{Dataset Curation}
\label{supp:curation}
To construct \ourbenchmark, we develop a hybrid automated and human-in-the-loop process that converts existing non-region-based 4D VQA benchmarks into region-based format.
Recall ~\secref{sec:benchmark} and Fig.~\ref{fig:region4d}, our curation process consists of the following stages.

\myparagraph{(a) Keyord Extraction.}
Given a question $\mQ$ and the first frame $\mI^{(1)}$ of a video, we first identify the objects mentioned in $\mQ$.
We employ Qwen2.5-VL-32B-Instruct~\cite{alibab2025qwen25vl} to parse the question and extract object references.
The model is given the following system prompt.
\begin{tcolorbox}[colback=gray!10, colframe=gray!50, boxrule=0.5pt, arc=2pt, breakable]
\small
\textbf{Task:} You will receive (1) an RGB image (the first frame of a video) and (2) a natural-language question about objects in the image.

\textbf{Instructions:} Identify the object(s) mentioned in the question and wrap them with angle brackets \texttt{<>}. Do not change any other part of the text. If no object matches, return the original question.

\textbf{Example:}\\
\textit{Input:} ``What is the teacher right hand holding?''\\
\textit{Output:} ``What is the \texttt{<teacher>} right hand holding?''
\end{tcolorbox}

\myparagraph{(b) Detect \& Segment.}
If the segmentation masks of the identified objects are annotated in the original source, \eg, DAVIS~\cite{perazzi2016davis,ponttuset2017davis}, we skip this stage.
Otherwise, we extract the 2D bounding boxes and segmentation masks for each identified object using a combination of GroundingDINO~\cite{liu2024groundingdino} and SAM2~\cite{ravi2024sam2}.
Specifically, we use GrondingDINO (\inlinecode{IDEA-Research/grounding-dino-base} from {\tt HuggingFace}) to detect objects based on the extracted object classes from (a).
We set both detection and text thresholds to 0.25.
The detected bounding boxes are then refined using SAM2 (\inlinecode{sam2.1\_hiera\_large}) to obtain refined segmentation masks.

\myparagraph{(c) Set of Marks.}
We leverage Set-of-Mark (SoM)~\cite{yang2023som} to generate an intermediate region-based visual, serving as a bridge to convert non-region-based inputs into our final region-based format.
We overlay numbered markers on the detected objects in $\mI^{(1)}$, creating an annotated image where each object is labeled with a unique ID and its class name, \eg, ``0:cat'', ``1:table''.
An example image is shown in~\figref{fig:r4d_som}.

\input{figures/supp/r4d_som}

\myparagraph{(d) Matching.}
We feed the annotated image from (c) and $\mQ$ into Qwen2.5-VL-32B-Instruct with the following prompt to match the objects in $\mQ$ to the marked regions.

\begin{tcolorbox}[colback=gray!10, colframe=gray!50, boxrule=0.5pt, arc=2pt, breakable,
                  top=3pt, bottom=3pt, left=4pt, right=4pt]
\small
\textbf{Task:} You will receive (1) an RGB image with labeled objects (a frame from a video) and (2) a natural-language question.

\textbf{Instructions:}
\begin{itemize}[leftmargin=12pt, topsep=2pt, itemsep=2pt]
    \item Identify which labeled objects the question refers to
    \item Replace object mentions with tokens: \texttt{<obj\_1>}, \texttt{<obj\_2>}, etc.
    \item If no objects match, return the original question with empty \texttt{obj\_classes}
\end{itemize}

\textbf{Output Format:} End your answer with ``\texttt{\#\#\# Final Answer:}'' followed by JSON:
\begin{verbatim}
{
  "question": "...",
  "obj_classes": ["id:class_name", ...]
}
\end{verbatim}

\textbf{Examples:}

\textit{Q:} ``What is the color of the car?''\\
(car labeled as \texttt{1:car})\\
\textit{A:}
\begin{verbatim}
{
  "question": "What is the color of
               <obj_1>?",
  "obj_classes": ["1:car"]
}
\end{verbatim}

\textit{Q:} ``What is the color of the cars?''\\
(two cars: \texttt{1:car}, \texttt{2:car})\\
\textit{A:}
\begin{verbatim}
{
  "question": "What is the color of
               <obj_1> and <obj_2>?",
  "obj_classes": ["1:car", "2:car"]
}
\end{verbatim}

\textit{Q:} ``What is the color of the car?''\\
(no car labeled)\\
\textit{A:}
\begin{verbatim}
{
  "question": "What is the color of
               the car?",
  "obj_classes": []
}
\end{verbatim}
\end{tcolorbox}

\myparagraph{(e) Verification.}
We manually verify all converted questions to ensure quality.
We use Label Studio~\cite{labelstudio} to build a simple interface where human annotators can review each QA pair along with the video and the detected regions.
Questions where the grounding fails, \ie, no objects detected or object misalignment, are fixed by annotators.
If a question cannot be fixed, it is filtered out.
We trim down the input video if the object appears later in the video instead of the first frame.
We exclude VQA sample where the object of interest in $\mQ$ is too ambiguous to ground clearly for our human annotators.
Overall, samples requiring correction or removal account for more than 50\% of the candidate samples from the automated pipeline, underscoring the necessity of this human verification step.
The final \ourbenchmark~contains 1,419 region-based QA pairs.

%% file: figures/supp/r4d_som.tex
\begin{figure}[t]
    \centering
    \includegraphics[width=\linewidth]{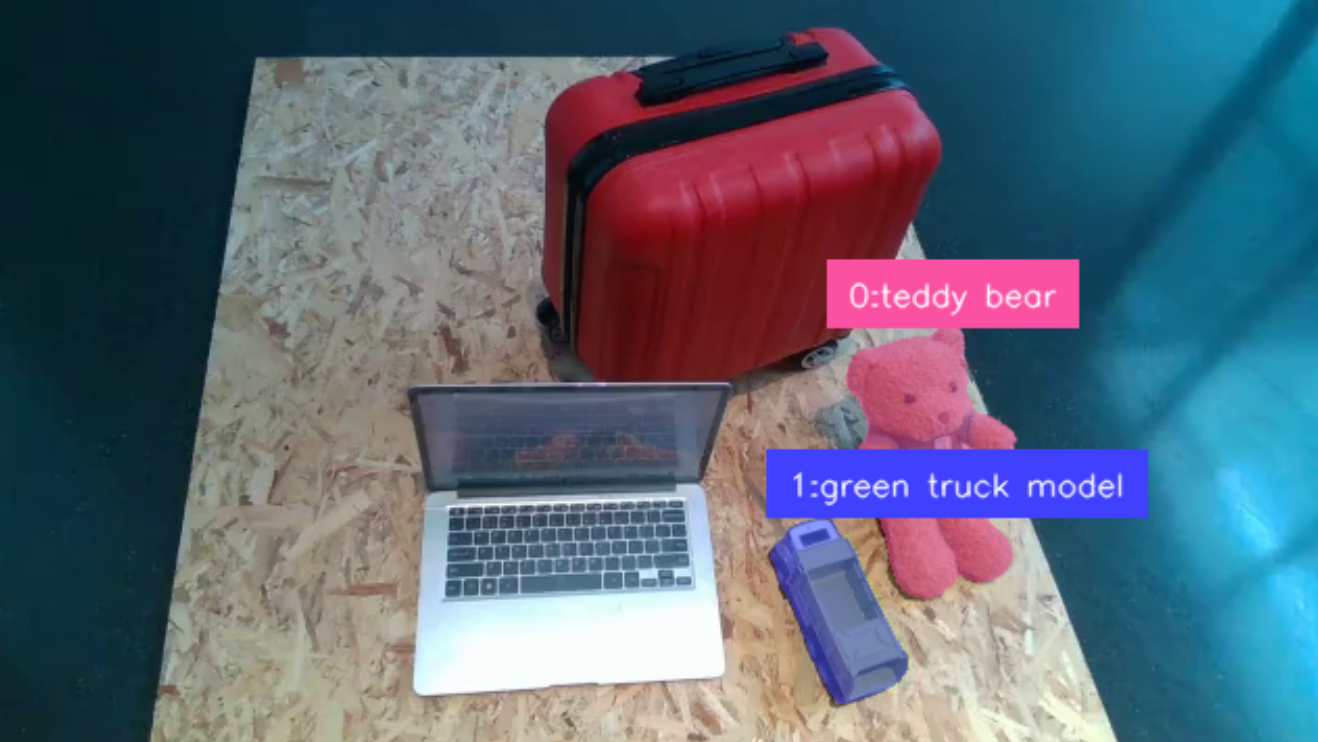}
    \caption{
        \textbf{An example of SoM visual input in \ourbenchmark.}
        We apply SoM~\cite{yang2023som} on $\mI^{(1)}$ to generate intermediate region-based visual inputs.
        The corresponding input $\mQ$ is ``\textit{At 9.00 sec, what is the positional relationship of the \textcolor{blue}{green truck model} relative to the \textcolor{brinkpink}{teddy bear}?}''
    \label{fig:r4d_som}}
\end{figure}

%% file: sections/supp/r4d_category.tex
\subsection{Question Categories}
\label{supp:categories}
\ourbenchmark~contains 9 question categories covering both \tcbox[lightgraybox]{static} and \tcbox[lightpurplebox]{dynamic} aspects of 4D understanding.
Of the 9 categories, 4 of them are sourced from VLM4D~\cite{zhou2025vlm4d} and the other 5 are sourced from STI-Bench~\cite{li2025stibench}.
For each category, we provide its defintiion below. We also attach several video examples in the supplementary folder under \inlinecode{r4d\_examples/}.

For the \textbf{T}ranslational (\tcbox[lightpurplebox]{T}), \textbf{R}otational (\tcbox[lightpurplebox]{R}), \textbf{C}ounting (\tcbox[lightpurplebox]{C}), and \textbf{F}alse \textbf{P}ositive (\tcbox[lightpurplebox]{FP}) questions, we follow the definitions in VLM4D~\cite{zhou2025vlm4d}.
We downloaded the dataset from their official source on HuggingFace, \ie, \inlinecode{shijiezhou/VLM4D}.
However, as of the time of writing, they do not provide the list of QA pairs for each category.
Therefore, we leverage Qwen2.5-VL-32B-Instruct~\cite{alibab2025qwen25vl} and human annotators to classify each QA pair into the 4 categories.
Of the region-based QA pairs in \ourbenchmark~obtained from VLM4D, the distribution across different categories is as follows:
\begin{itemize}
    \item Translational: 61.3\%
    \item Rotational: 10.2\%
    \item Counting: 15.4\%
    \item False Positive: 13.1\%
\end{itemize}
In comparison, the official VLM4D benchmark has the following distribution:
\begin{itemize}
    \item Translational: 55\%
    \item Rotational: 19\%
    \item Counting:  17\%
    \item False Positive: 9\%
\end{itemize}
Our categorization results are largely consistent with the official distribution with slight difference.

For the 3D \textbf{V}ideo \textbf{G}rounding (\tcbox[lightgraybox]{VG}), \textbf{S}patial \textbf{R}elationship (\tcbox[lightgraybox]{SR}), \textbf{D}imension \textbf{M}easurement (\tcbox[lightgraybox]{DM}), \textbf{D}isplacement \& \textbf{P}ath Length (\tcbox[lightpurplebox]{DP}), and \textbf{S}peed \& \textbf{A}cceleration (\tcbox[lightpurplebox]{SA}) questions,
we follow the definition of STI-Bench~\cite{li2025stibench}.
We downloaded the dataset from their official source on HuggingFace, \ie, \inlinecode{MINT-SJTU/STI-Bench}.
We note that the original STI-Bench contains two additional categories, \ie, \textit{Ego-centric Orientation} and \textit{Trajectory Description},
where these questions focuses on the ego-centric 4D understanding from the viewpoint itself.
Since \ourbenchmark~focuses on region-based 4D VQA, where another region of interest needs to be provided, these questions are not applicable and removed from \ourbenchmark.

\input{figures/supp/r4d_translational}
\input{figures/supp/r4d_rotational}
\input{figures/supp/r4d_counting}
\input{figures/supp/r4d_false_positive}
\input{figures/supp/r4d_grounding}
\input{figures/supp/r4d_relation}
\input{figures/supp/r4d_measurement}
\input{figures/supp/r4d_displacement}
\input{figures/supp/r4d_speed}

The followings are the detailed explanations for each category:

\myparagraph{\textbf{T}ranslational (\tcbox[lightpurplebox]{T})} questions target the MLLM's capabilities to understand the linear movement of objects.
They usually involve the following movement-related diretion, such as left, right, north, south, away, towards, etc.
We provide several examples of \ourbenchmark~translational questions in Fig.~\ref{fig:r4d_translational}.

\myparagraph{\textbf{R}otational (\tcbox[lightpurplebox]{R})} questions, on the other hand, care about the rotational movement of objects.
They usually involve the following movement-related words, such as rotate, spin, twist, turn, etc.
We provide several examples of \ourbenchmark~rotational questions in Fig.~\ref{fig:r4d_rotational}.

\myparagraph{\textbf{C}ounting (\tcbox[lightpurplebox]{C})} questions focusing on the MLLM's ability to accurately count the number of objects or occurrences of actions.
We provide several examples of \ourbenchmark~counting questions in Fig.~\ref{fig:r4d_counting}.

\myparagraph{\textbf{F}alse \textbf{P}ositive (\tcbox[lightpurplebox]{FP})} questions are designed to trick the MLLM.
The questions will intentionally describe events that do not actually occur within the video,
\eg, asking about movements when no object is moving.
We note that the original VLM4D false positive questions also ask about objects that do not exist in the video.
Due to the nature of region-based 4D VQA in \ourbenchmark, we do not include these types of questions since the regions cannot refer to non-existent objects.
We provide several examples of \ourbenchmark~false positive questions in Fig.~\ref{fig:r4d_false_positive}.

\myparagraph{3D \textbf{V}ideo \textbf{G}rounding (\tcbox[lightgraybox]{VG})} questions ask MLLMs to retrive the 3D bounding box of objects.
The options are formatted as JSON with ``dimension (size)'' $\in \sR^3$, ``central point (coordinate)'' $\in \sR^3$ and ``orientation'' $\in \sR^3$, (\ie, yawn, pitch, and roll) or ``camera heading'' $\in \sR^1$.
We provide an example in Fig.~\ref{fig:r4d_grounding}.
As shown in the example, the MLLM needs to be fairly precise to answer these questions correctly, as the differences between options can be quite small.

\myparagraph{\textbf{S}patial \textbf{R}elationship (\tcbox[lightgraybox]{SR})} questions assess the 3D spatial relationship between selected objects or the camera.
The options usually involve relative positioning terms, such as left, right, front, back, up, down, etc.
We provide an example of \ourbenchmark~spatial relation questions in Fig.~\ref{fig:r4d_relation}.

\myparagraph{\textbf{D}imension \textbf{M}easurement (\tcbox[lightgraybox]{DM})} questions care about the physical measurements of objects, such as size and distance.
They usually require MLLMs to understand and perceive depth information in order to predict the numerical values.
We provide an example of \ourbenchmark~dimension measurement questions in Fig.~\ref{fig:r4d_measurement}.

\myparagraph{\textbf{D}isplacement \& \textbf{P}ath Length (\tcbox[lightpurplebox]{DP})} questions measures the travel distance of objects.
They often involve MLLMs to track motion across selected frames.
We provide an example of \ourbenchmark~displacement and path length questions in Fig.~\ref{fig:r4d_displacement}.

\myparagraph{\textbf{S}peed \& \textbf{A}cceleration (\tcbox[lightpurplebox]{SA})} questions estimate the motion dynamics of objects.
The MLLM needs to consider both the displacement and time intervals to answer them correctly.
We provide an example of \ourbenchmark~speed and acceleration questions in Fig.~\ref{fig:r4d_speed}.

%% file: figures/supp/r4d_translational.tex
\begin{figure}[t!]
    \centering
    \includegraphics[width=\linewidth]{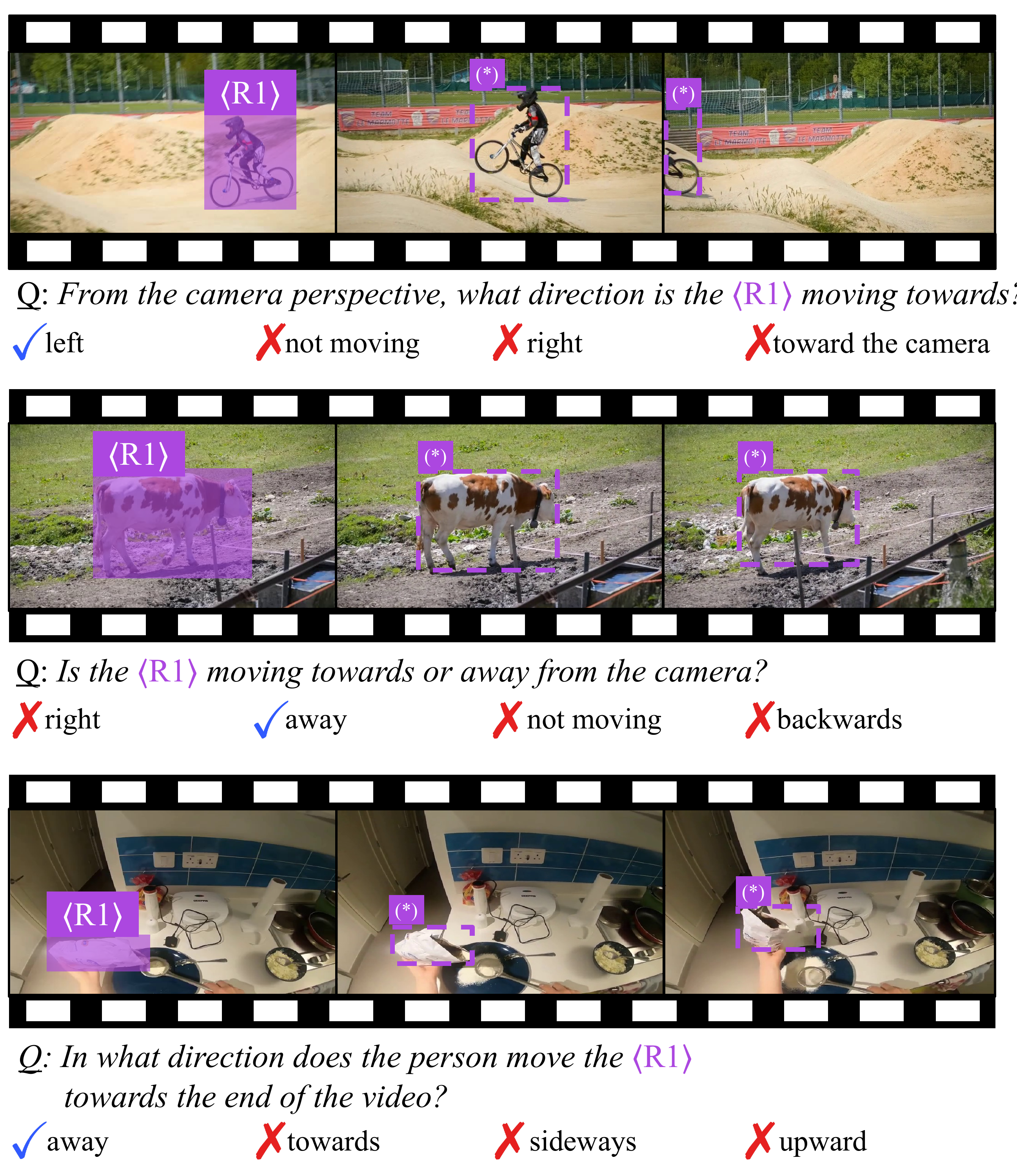}
    \caption{
        \textbf{Translational questions in \ourbenchmark.}
        We note that the regions labeled with \tcbox[tightpurple]{(*)} are not provided in \ourbenchmark; they are visualized for readability.
    }
    \label{fig:r4d_translational}
\end{figure}

%% file: figures/supp/r4d_rotational.tex
\begin{figure}[ht!]
    \centering
    \includegraphics[width=\linewidth]{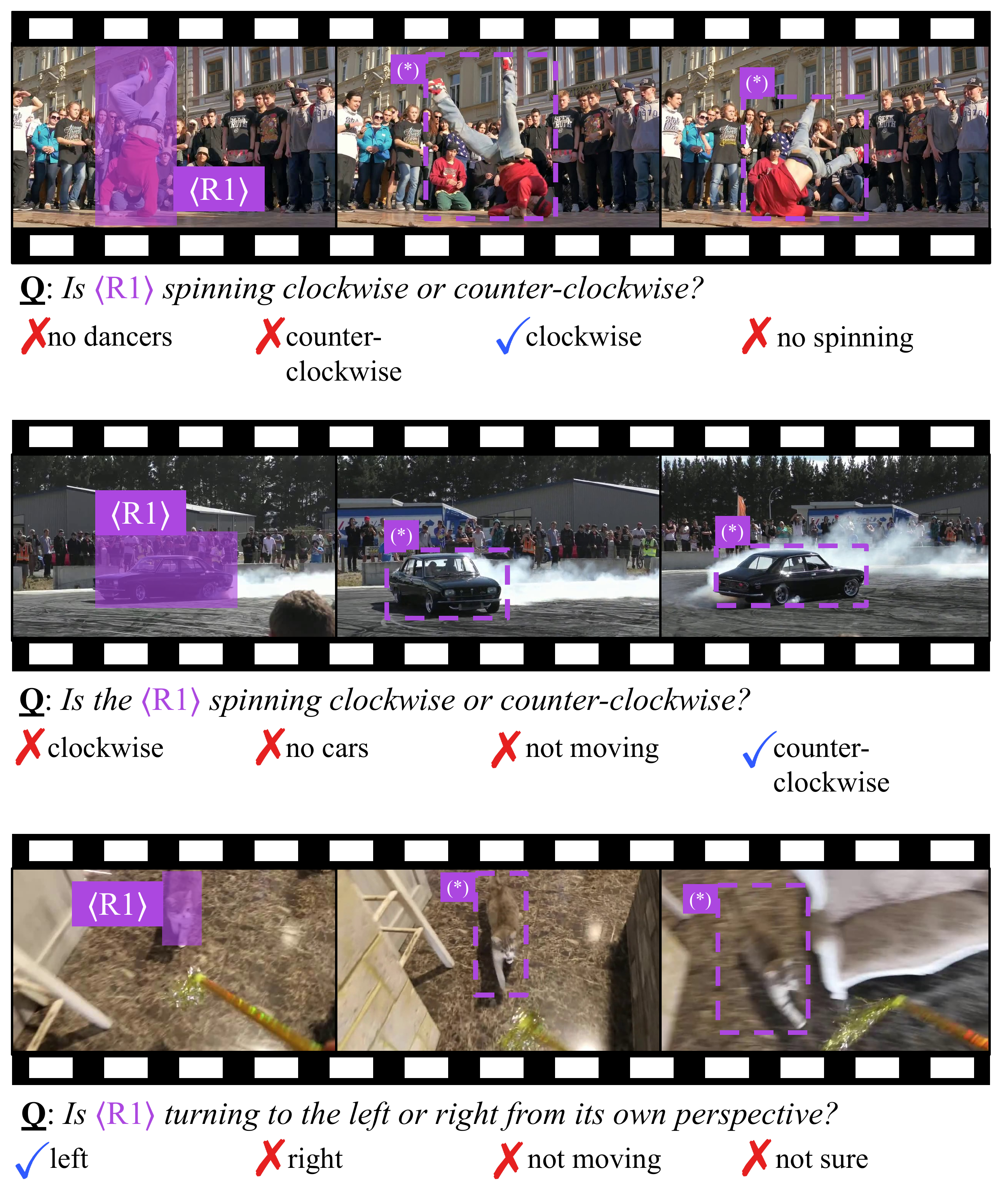}
    \caption{
        \textbf{Rotational questions in \ourbenchmark.}
        We note that the regions labeled with \tcbox[tightpurple]{(*)} are not provided in \ourbenchmark; they are visualized for readability.
    }
    \label{fig:r4d_rotational}
\end{figure}

%% file: figures/supp/r4d_counting.tex
\begin{figure}[ht!]
    \centering
    \includegraphics[width=\linewidth]{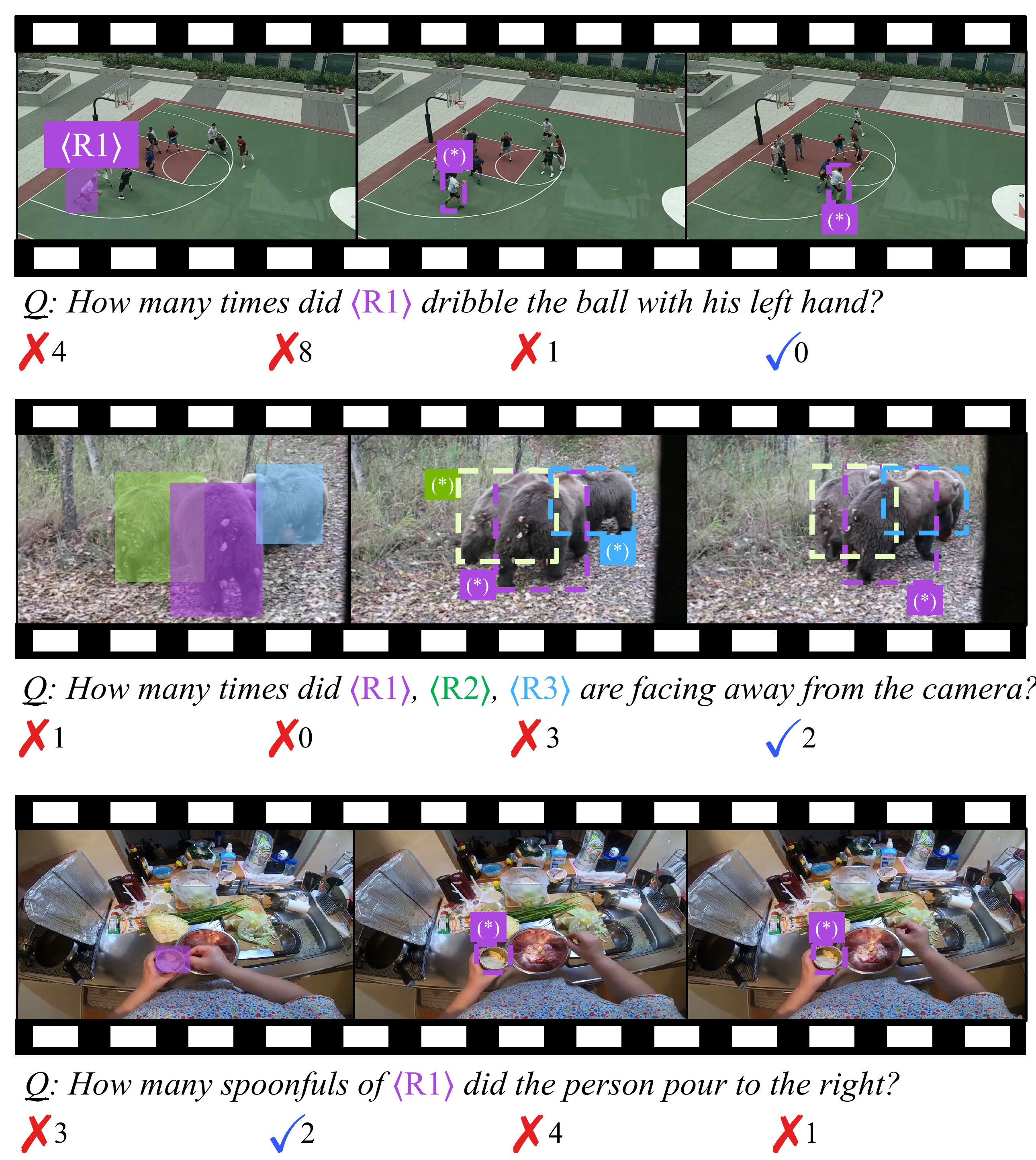}
    \caption{
        \textbf{Counting questions in \ourbenchmark.}
        We note that the regions labeled with \tcbox[tightpurple]{(*)}, \tcbox[tightgreen]{(*)}, or \tcbox[bluebox]{(*)} are not provided in \ourbenchmark; they are visualized for readability.
    }
    \label{fig:r4d_counting}
\end{figure}

%% file: figures/supp/r4d_false_positive.tex
\begin{figure}[ht!]
    \centering
    \includegraphics[width=\linewidth]{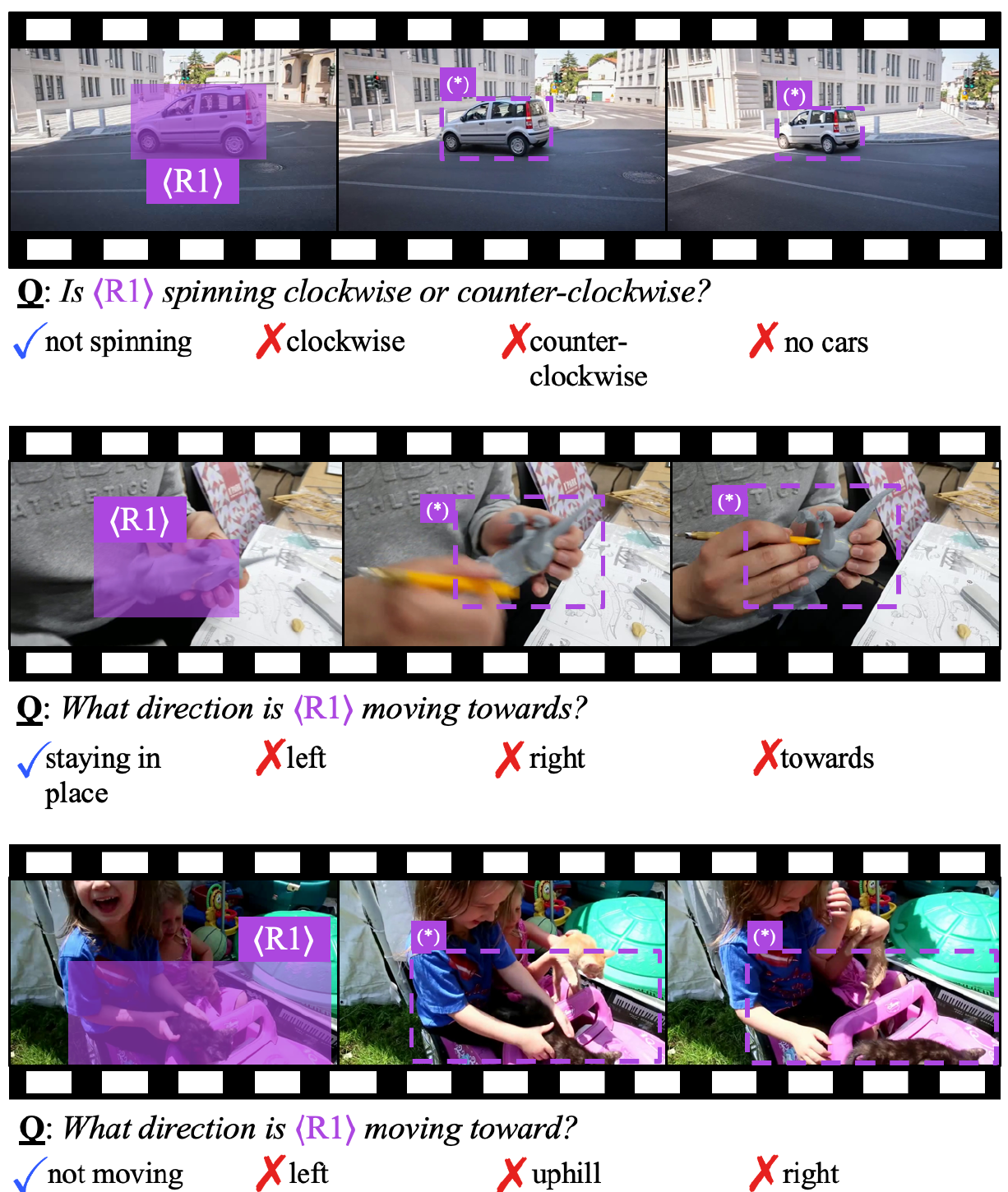}
    \caption{
        \textbf{False positive questions in \ourbenchmark.}
        We note that the regions labeled with \tcbox[tightpurple]{(*)} are not provided in \ourbenchmark; they are visualized for readability.
    }
    \label{fig:r4d_false_positive}
\end{figure}

%% file: figures/supp/r4d_grounding.tex
\begin{figure}[ht!]
    \centering
    \includegraphics[width=\linewidth]{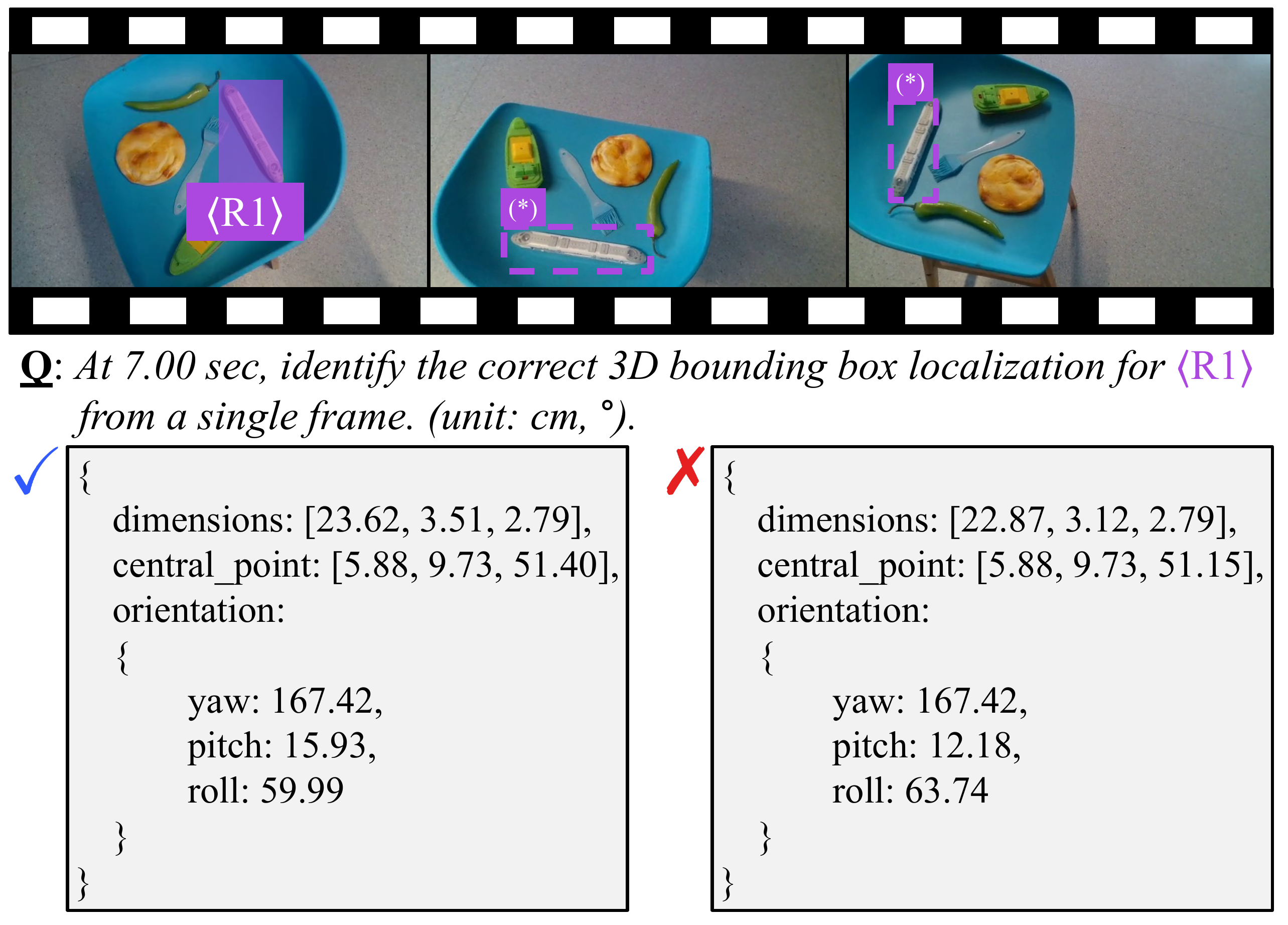}
    \caption{
        \textbf{3D video grounding questions in \ourbenchmark.}
        We note that the regions labeled with \tcbox[tightpurple]{(*)} are not provided in \ourbenchmark; they are visualized for readability.
        For simplicity, we only show 1 correct option and 1 wrong option here, but there are 5 options for each 3D video grounding question in \ourbenchmark.
    }
    \label{fig:r4d_grounding}
\end{figure}

%% file: figures/supp/r4d_relation.tex
\begin{figure}[ht!]
    \centering
    \includegraphics[width=\linewidth]{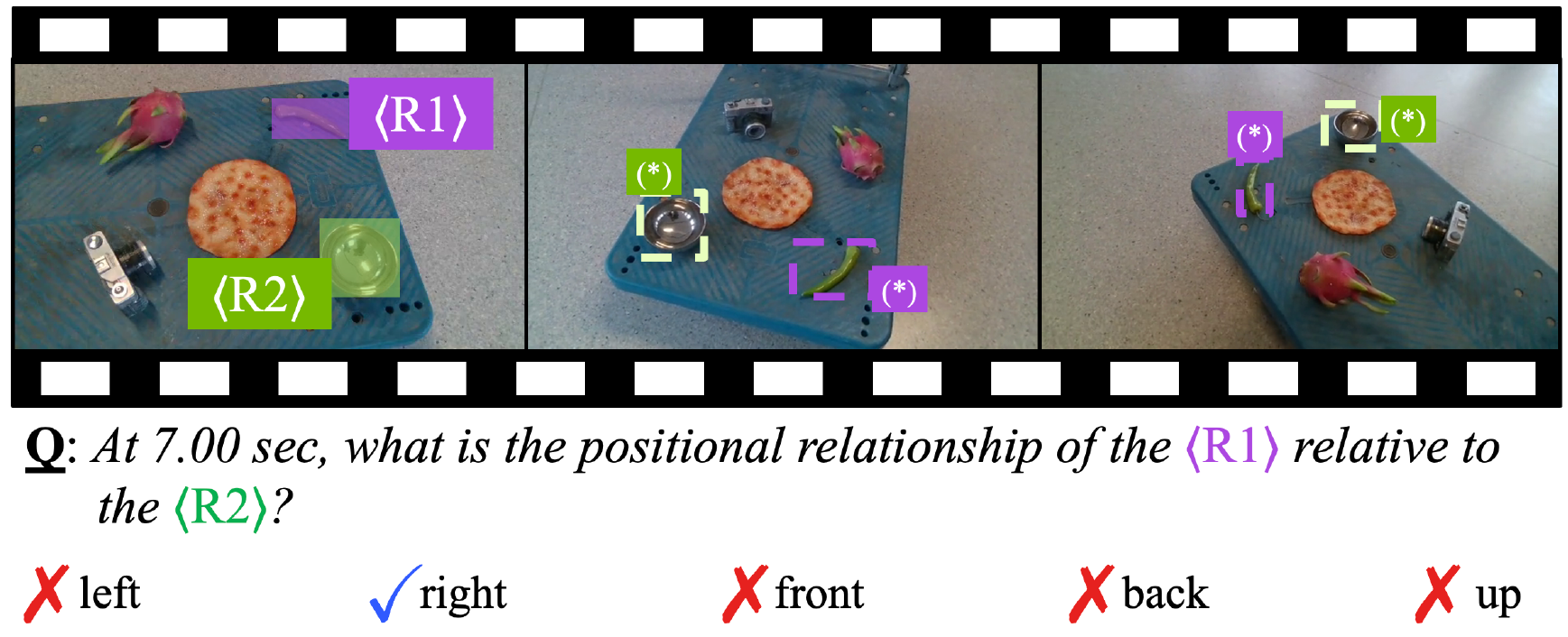}
    \caption{
        \textbf{Spatial relation questions in \ourbenchmark.}
        The question asks about the spatial relationship at 7 seconds, which corresponds to the middle frame out of the three frames shown.
        We note that the regions labeled with \tcbox[tightpurple]{(*)} or \tcbox[tightgreen]{(*)} are not provided in \ourbenchmark; they are visualized for readability.
    }
    \label{fig:r4d_relation}
\end{figure}

%% file: figures/supp/r4d_measurement.tex
\begin{figure}[ht!]
    \centering
    \includegraphics[width=\linewidth]{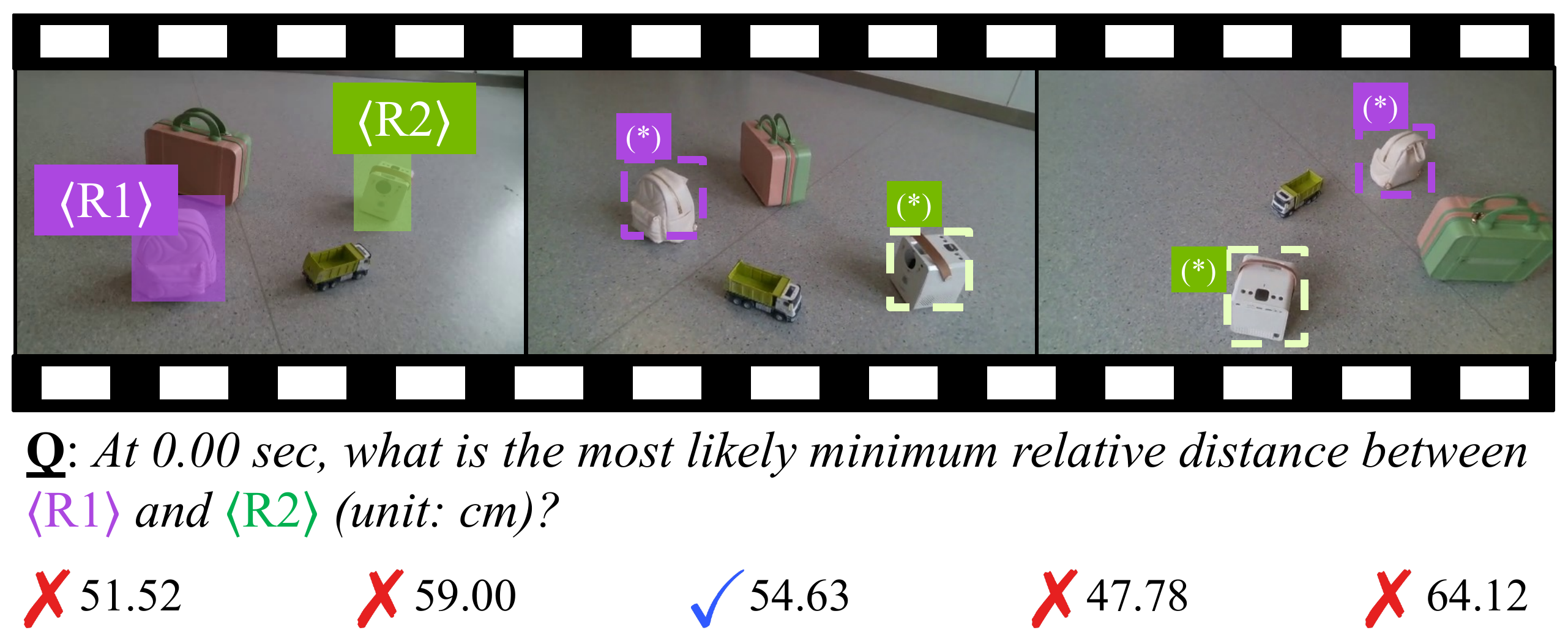}
    \caption{
        \textbf{Dimension measurement questions in \ourbenchmark.}
        We note that the regions labeled with \tcbox[tightpurple]{(*)} or \tcbox[tightgreen]{(*)} are not provided in \ourbenchmark; they are visualized for readability.
    }
    \label{fig:r4d_measurement}
\end{figure}

%% file: figures/supp/r4d_displacement.tex
\begin{figure}[ht!]
    \centering
    \includegraphics[width=\linewidth]{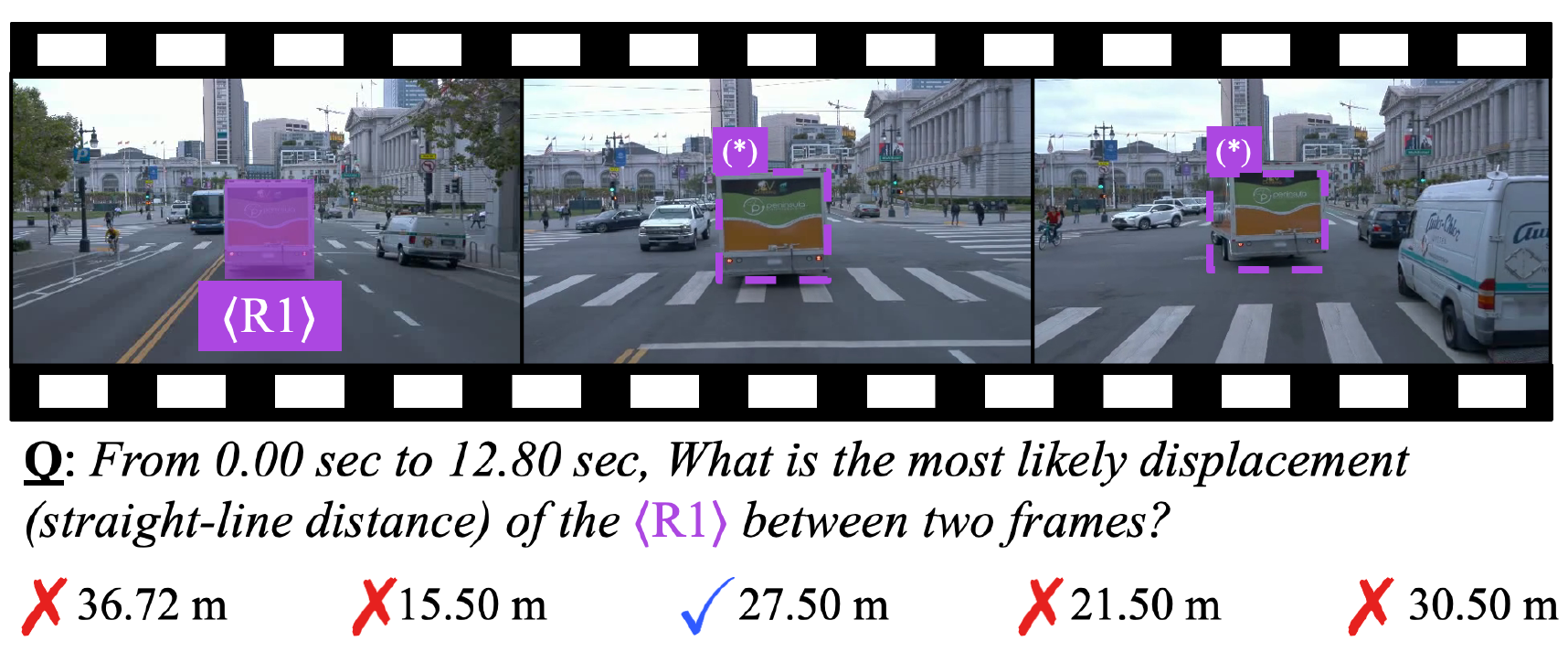}
    \caption{
        \textbf{Displacement \& path length questions in \ourbenchmark.}
        We note that the regions labeled with \tcbox[tightpurple]{(*)} are not provided in \ourbenchmark; they are visualized for readability.
    }
    \label{fig:r4d_displacement}
\end{figure}

%% file: figures/supp/r4d_speed.tex
\begin{figure}[ht!]
    \centering
    \includegraphics[width=\linewidth]{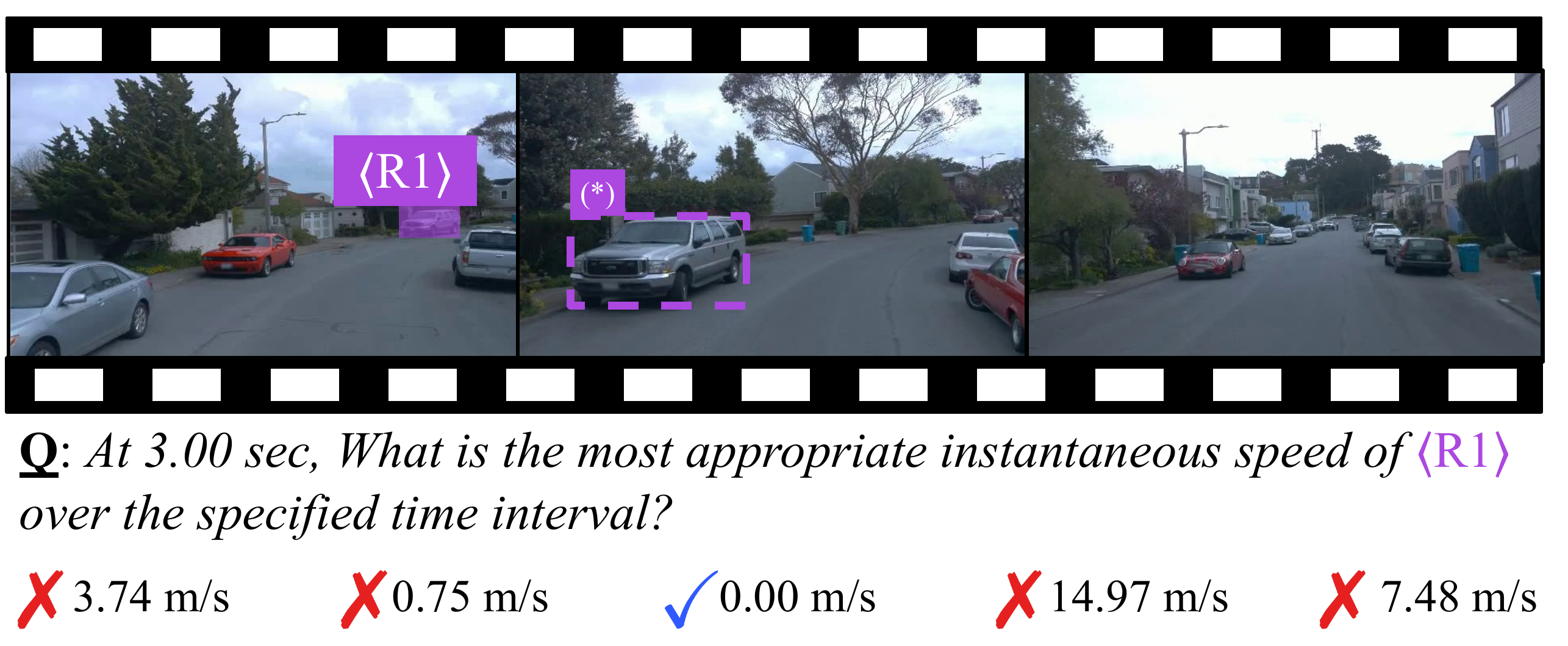}
    \caption{
        \textbf{Speed \& acceleration questions in \ourbenchmark.}
        We note that the regions labeled with \tcbox[tightpurple]{(*)} are not provided in \ourbenchmark; they are visualized for readability.
    }
    \label{fig:r4d_speed}
\end{figure}

%% file: tables/supp/more_nvila_nonregion.tex
\centering
\captionof{table}{
    \textbf{Evaluation on non-region-level 3D / 4D benchmarks.}
    We report the average multiple-choice accuracy $(\uparrow)$ on each benchmark.
    For simplicity, we use the following abbreviations:
    STI (STI-Bench~\cite{li2025stibench}),
    V4D (VLM4D-real~\cite{zhou2025vlm4d}),
    MMSI (MMSI-Bench~\cite{yang2025mmsi}),
    OS (OmniSpatial~\cite{jia2025omnispatial}),
    and
    VSTI (VSTI-Bench~\cite{fan2025vstibench}).
    \label{tab:more_nvila_nonregion}
}
\setlength{\tabcolsep}{3pt}
\resizebox{0.85\linewidth}{!}{
\begin{tabular}{l ccc ccc c}
    \specialrule{.15em}{.05em}{.05em}
    {Methods}
    & STI
    & V4D
    & MMSI
    & OS
    & SAT
    & VSTI
    \\
    \midrule
    NVILA-Lite-8B & 33.8 & 46.5 & 31.3 & 37.2 &  62.0 & 45.2
    \\
    \oursrow
    & 37.6 & 52.7 & 33.3 & 40.4 & 64.7 & 59.1
    \\
    \noalign{\vskip-0.5pt} 
    \oursrow
    \multirow{-2}{*}{\oursabbr-8B (Ours)}
    & \improve{+3.8} & \improve{+6.2} & \improve{+2.0} & \improve{+3.2} & \improve{+2.7} & \improve{+13.9}
    \\
    NVILA-Lite-15B
    & 34.2 & 45.1  & 29.5 & 41.0 & 62.7 & 42.4
    \\
    \oursrow
    &
    38.1 & 53.7  & 31.7 & 42.7 & 65.3 & 58.6
    \\
    \noalign{\vskip-0.5pt}
    \oursrow
    \multirow{-2}{*}{\oursabbr-15B (Ours)}
    & \improve{+3.9} & \improve{+8.6} & \improve{+2.2} & \improve{+1.7} & \improve{+2.6} & \improve{+16.2}
    \\
    \specialrule{.15em}{.05em}{.05em}
\end{tabular}
}

%% file: tables/supp/more_nvila_region4d.tex
\centering
\captionof{table}{
    \textbf{Evaluation on \ourbenchmark.}
    We report performance on the static split (\tcbox[lightgraybox]{\bf Sta}), the dynamic split (\tcbox[lightpurplebox]{\bf Dyn}), and all 9 tasks of \ourbenchmark.
    For simplicity, we abbreviate them as follows:
    3D \textbf{V}ideo \textbf{G}rounding (\tcbox[lightgraybox]{VG});
    \textbf{D}imension \textbf{M}easurement (\tcbox[lightgraybox]{DM});
    \textbf{S}patial \textbf{R}elationship (\tcbox[lightgraybox]{SR});
    \textbf{R}otational (\tcbox[lightpurplebox]{R});
    \textbf{C}ounting (\tcbox[lightpurplebox]{C});
    \textbf{T}ranslational (\tcbox[lightpurplebox]{T});
    \textbf{F}alse \textbf{P}ositive (\tcbox[lightpurplebox]{FP});
    \textbf{S}peed \& \textbf{A}cceleration (\tcbox[lightpurplebox]{SA});
    and
    \textbf{D}isplacement \& \textbf{P}ath Length (\tcbox[lightpurplebox]{DP}).
    \label{tab:more_nvila_region4d}
}
\small
\setlength{\tabcolsep}{2pt}
\resizebox{\linewidth}{!}{
\begin{tabular}{lc<{\hskip 8pt}c c c<{\hskip 8pt} c cc cccccccc}
    \specialrule{.15em}{.05em}{.05em}
    {Methods}
    & {\bf Avg}
    & \cellcolor{lightgray}{\bf Sta}   & \cellcolor{lightpurple}{\bf Dyn} &
    & \cellcolor{lightgray}{VG}   & \cellcolor{lightgray}{DM}     & \cellcolor{lightgray}{SR} &
    & \cellcolor{lightpurple}{R}    & \cellcolor{lightpurple}{C}    & \cellcolor{lightpurple}{T} & \cellcolor{lightpurple}{FP}
    & \cellcolor{lightpurple}{SA}   & \cellcolor{lightpurple}{DP}
    \\
    \midrule
    NVILA-Lite-8B
    & 37.9  & 29.1     & 41.3 &
    & 33.9 & 20.2 & 46.3 &
    & 41.5 & 39.6 & 41.9 & 40.7
    & 45.9 & 32.1
    \\
    \oursrow
    & 42.2
    & 32.9 & 45.7 &
    & 35.1 & 26.3 & 52.2 &
    & 43.1 & 40.1 & 48.7 & 40.2
    & 50.9 & 38.9
    \\
    \noalign{\vskip-0.5pt}
    \oursrow
    \multirow{-2}{*}{\oursabbr-8B (Ours)}
    & \improve{+4.3} & \improve{+3.8} & \improve{+4.4} &
    & \improve{+1.2} & \improve{+6.1} & \improve{+5.9} &
    & \improve{+1.6} & \improve{+0.5} & \improve{+6.8} & -0.5
    & \improve{+5.0} & \improve{+6.8}
    \\
    \midrule
    NVILA-Lite-15B
    & 39.7  & 31.7 & 42.7 &
    & 36.5 & 26.8 & 31.7 &
    & 50.9 & 34.0 & 46.4 & 34.8 & 37.8 & 21.4
    \\
    \oursrow
    & 43.0 & 35.8 & 45.7 &
    & 38.5 & 32.2 & 39.0 &
    & 50.0 & 38.4 & 49.6 & 36.3 & 45.9 & 28.6
    \\
    \noalign{\vskip-0.5pt}
    \oursrow
    \multirow{-2}{*}{\oursabbr-15B (Ours)}
    & \improve{+3.3} & \improve{+4.1} & \improve{+3.10} &
    & \improve{+2.0} & \improve{+5.4} & \improve{+7.3} &
    & -0.9 & \improve{+4.4} & \improve{+3.2} & \improve{+1.5}
    & \improve{+7.9} & \improve{+7.2}
    \\
    \specialrule{.15em}{.05em}{.05em}
\end{tabular}
}

%% file: figures/timebench.tex
\centering
\includegraphics[width=\linewidth]{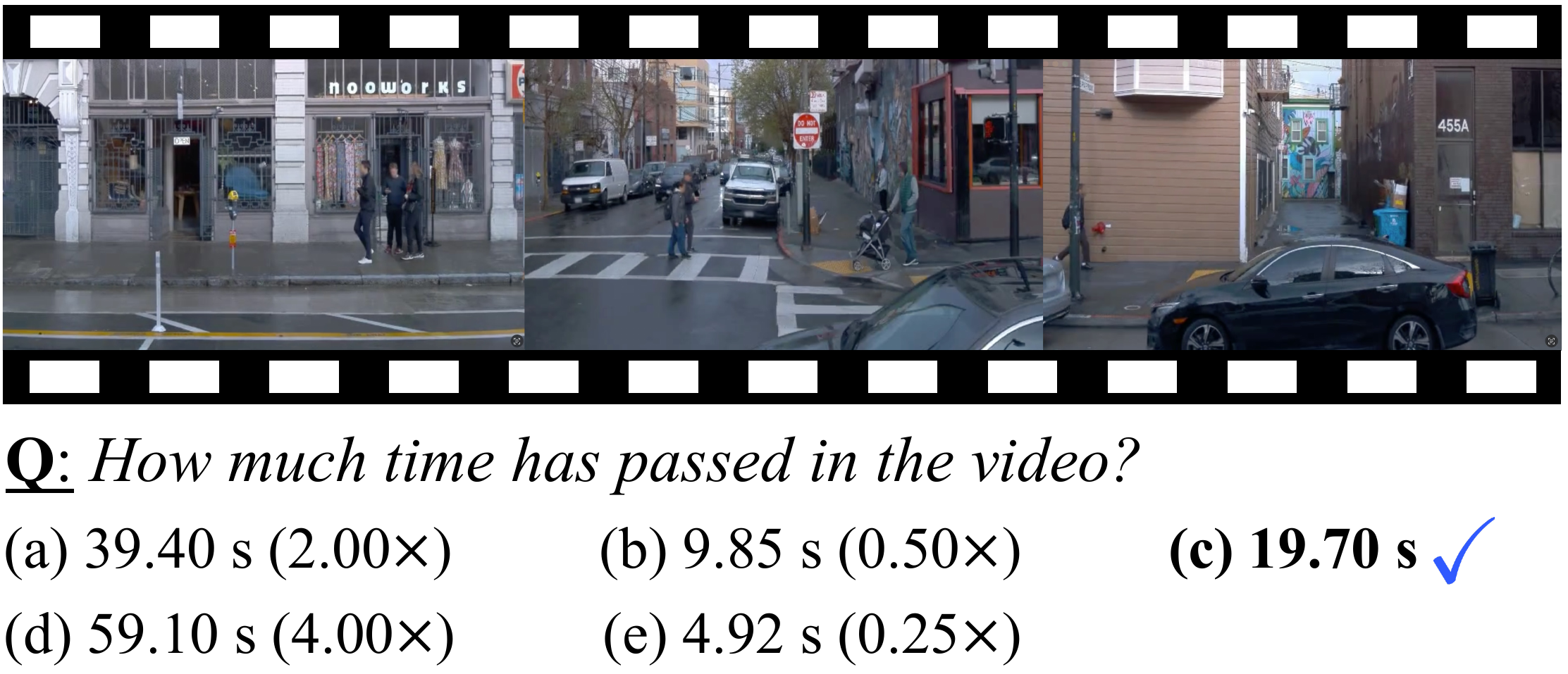} 
\caption{
    \textbf{TimeBench VQA.}
    We curate a toy benchmark to evaluate MLLMs' temporal perception.
    We note that the ``($M \times$)'' indicates the multiplier between the wrong option and the correct one.
    They are not provided in the actual question but are shown here for clarity.
}

%% file: tables/supp/timebench.tex
\centering
\captionof{table}{
    \textbf{Ablation studies on explicit temporal cues.}
    We experiment without and with different choices of explicit time cues.
    For simplicity, we use the same abbreviations as Tab.~\ref{tab:ablation_naive}.
    \label{tab:supp_timebench}
}
\resizebox{0.85\linewidth}{!}{
\begin{tabular}{l cccccc ccc c}
    \specialrule{.15em}{.05em}{.05em}
    {Methods} &
    {Time cues} &
    {TimeBench} &
    STI &
    {\ourbenchmarkabbr}
    \\
    \midrule
    {\it Zero-shot}
    & \ccross
    & 22.7
    & 33.8
    & 37.9
    \\
    \midrule
    {\it \ours} & \ccross
    & 30.1
    & 34.8
    & 41.0
    \\
    {\it \ours+mark}
    & marks
    & 95.3
    & 35.1
    & 41.1
    \\
    {\it \ours+prompt}
    & prompts
    & \bf 98.0
    & \bf 36.1
    & \bf 41.5
    \\
    \specialrule{.15em}{.05em}{.05em}
\end{tabular}
}
\vspace{-0.1cm}

%% file: tables/supp/dataset_increment.tex
\centering
\captionof{table}{
    \textbf{Incremental training data mixture.}
    We use the same abbreviations as Tab.~\ref{tab:ablation_naive} and the following for each dataset:
    \textbf{V}STI-Bench~\cite{fan2025vstibench} (V);
    \textbf{W}olf~\cite{li2024wolf} (W);
    \textbf{R}oboFAC~\cite{lu2025robofac} (R); 
    \textbf{S}AT~\cite{ray2024sat} (S).
}
\resizebox{\linewidth}{!}{
\begin{tabular}{l ccccc c ccc}
    \specialrule{.15em}{.05em}{.05em}
    \multirow{2}{*}{Methods} &
    \multirow{2}{*}{V} &
    \multirow{2}{*}{W} &
    \multirow{2}{*}{R} &
    \multirow{2}{*}{S} &
    \multirow{2}{*}{STI} &
    \multicolumn{3}{c}{\ourbenchmark}
    \\
    \cmidrule(lr){7-9}
    &
    & &&&
    & Avg
    & Sta
    & Dyn
    \\
    \midrule
    {\it Zero-shot}
    & \ccross & \ccross & \ccross & \ccross
    & 33.8 & 37.9 & 29.1 & 41.3
    \\
    \midrule
    {\it V}
    & \ccheck & \ccross & \ccross & \ccross
    & 35.4 & 39.4 & 30.0 & 42.9
    \\
    {\it V+W}
    & \ccheck & \ccheck & \ccross & \ccross
    & 36.0 & 40.6 & 31.0 & 44.2
    \\
    {\it V+W+R}
    & \ccheck & \ccheck & \ccheck & \ccross
    & 37.0 & 41.8 & 32.2 & 45.4
    \\
    \oursrow {\it V+W+R+S} (Ours)
    & \ccheck & \ccheck & \ccheck & \ccheck
    & \bf 37.6 & \bf 42.2 & \bf 32.9 & \bf 45.7
    \\
    \specialrule{.15em}{.05em}{.05em}
\end{tabular}
\label{tab:increment}
}

%% file: figures/supp/more_qual.tex
\begin{figure}[t]
    \centering
    \includegraphics[width=\linewidth]{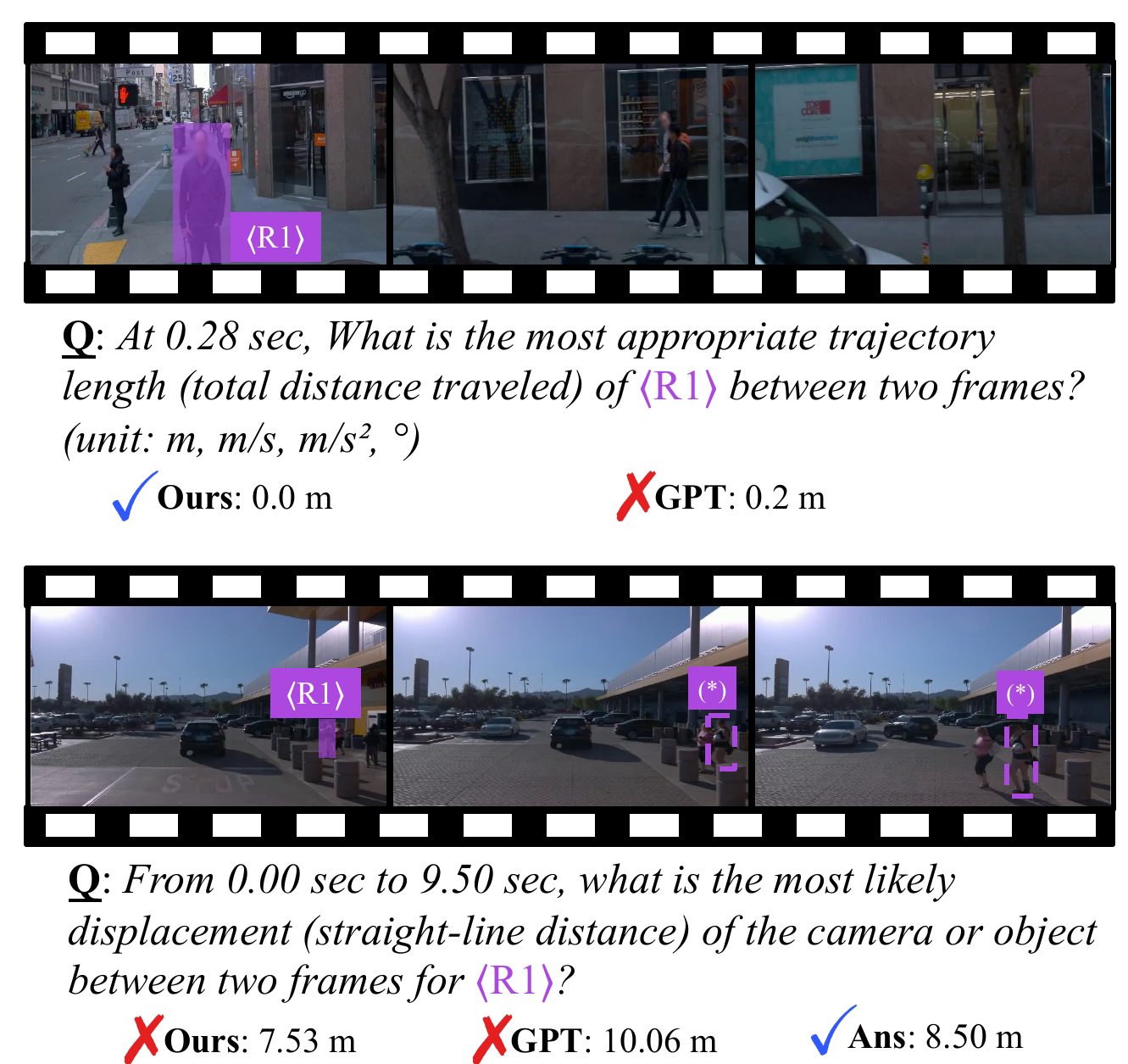}
    \caption{
        \textbf{More VQA comparison between GPT-4o~\cite{openai2024gpt4o} and \oursabbr~(Ours) on \ourbenchmark.}
        We provide 2 examples for each of the following categories: Displacement \& Path Length.
    }
    \label{fig:more_qual_4}
\end{figure}

\begin{figure}[ht]
    \centering
    \includegraphics[width=\linewidth]{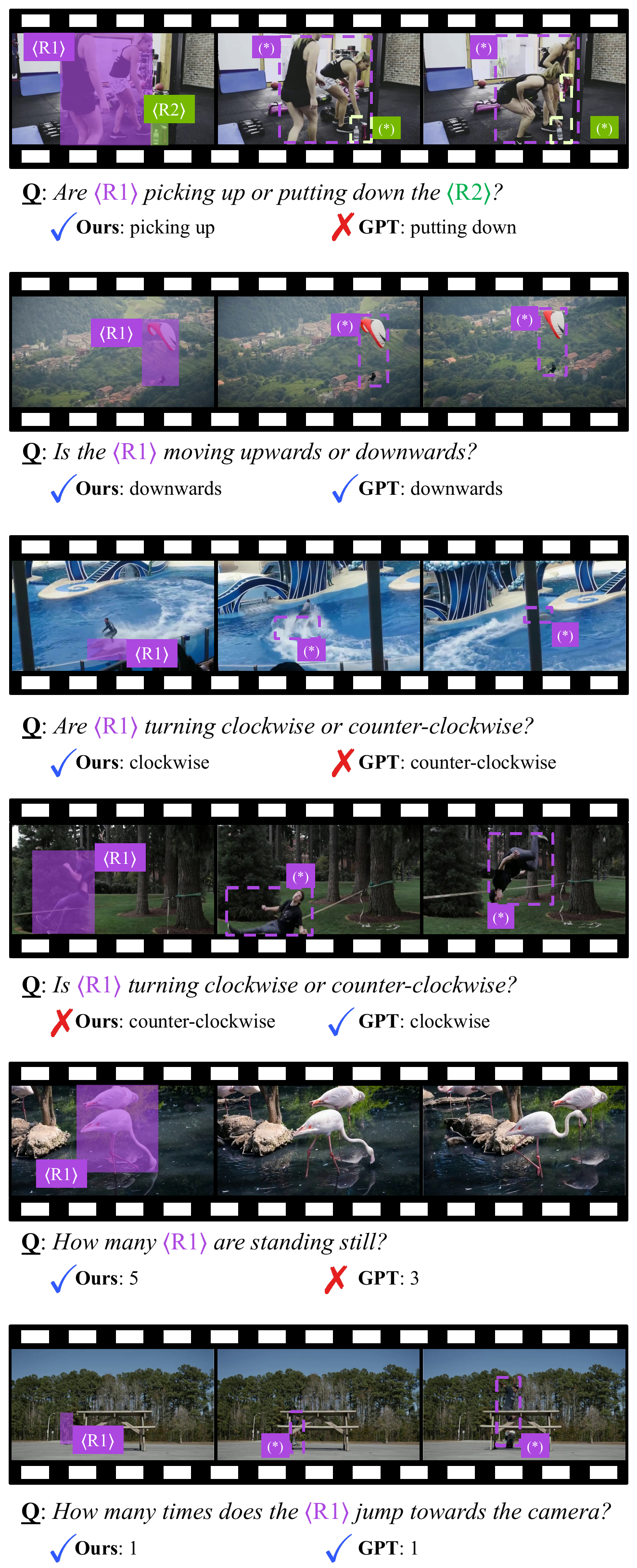}
    \caption{
        \textbf{More VQA comparison between GPT-4o~\cite{openai2024gpt4o} and \oursabbr~(Ours) on \ourbenchmark.}
        We provide 2 examples for each of the following categories: Translational, Rotational, and Counting.
    }
    \label{fig:more_qual_1}
\end{figure}

\begin{figure}[ht]
    \centering
    \includegraphics[width=\linewidth]{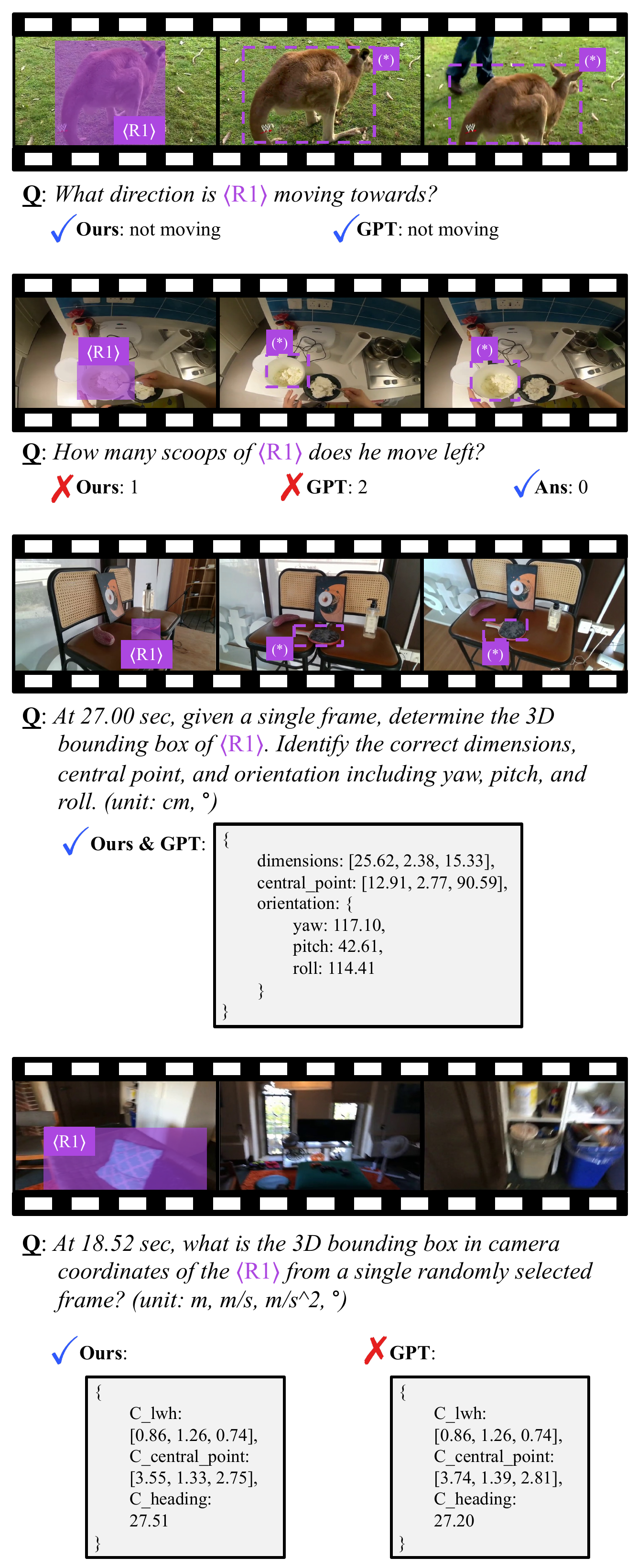}
    \caption{
        \textbf{More VQA comparison between GPT-4o~\cite{openai2024gpt4o} and \oursabbr~(Ours) on \ourbenchmark.}
        We provide 2 examples for each of the following categories: False Positive and 3D Video Grounding.
    }
    \label{fig:more_qual_2}
\end{figure}

\begin{figure}[ht]
    \centering
    \includegraphics[width=0.95\linewidth]{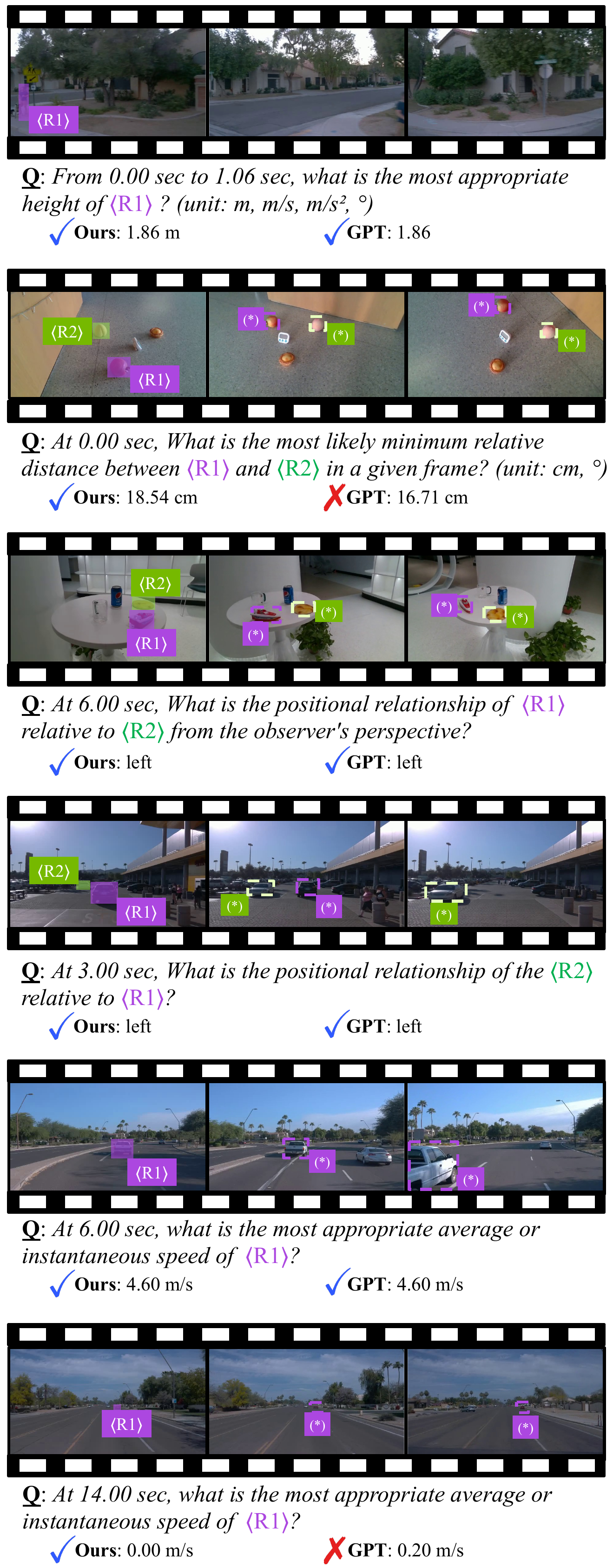}
    \caption{
        \textbf{More VQA comparison between GPT-4o~\cite{openai2024gpt4o} and \oursabbr~(Ours) on \ourbenchmark.}
        We provide 2 examples for each of the following categories: Spatial Relation, Dimension Measurement, and Speed \& Acceleration.
    }
    \label{fig:more_qual_3}
\end{figure}

%% file: figures/supp/more_depth_vis.tex
\begin{figure*}[t]
    \centering
    \includegraphics[width=\linewidth]{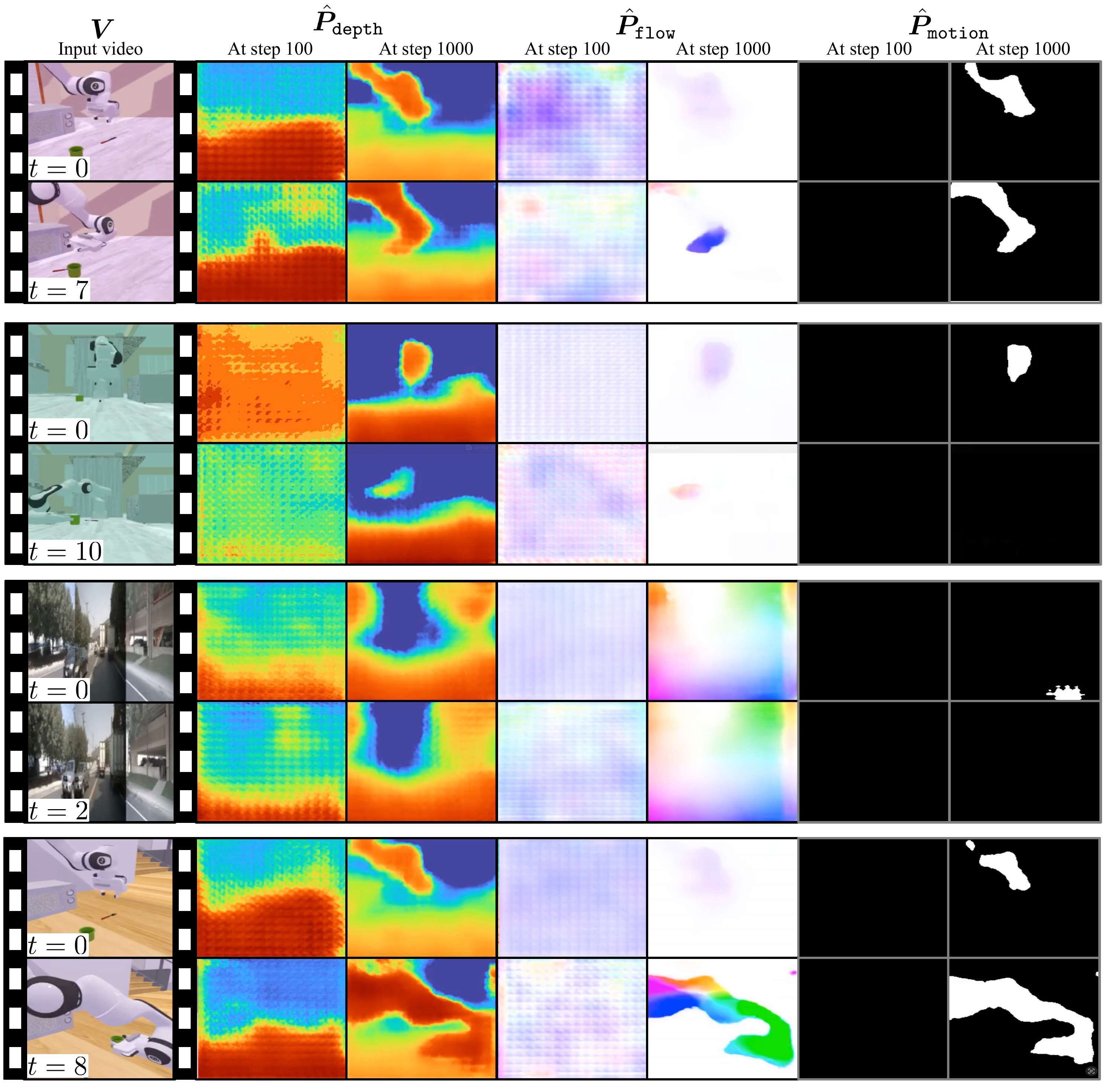}
    \caption{
        \textbf{More visualizations of \oursabbr~explicit signals $\hat \mP_{m}$.}
        Similar to the format of Fig.~\ref{fig:4DD_visual},
        we visualize the training progress of $\hat \mP_{\tt depth}$, $\hat \mP_{\tt flow}$, and $\hat \mP_{\tt motion}$.
    }
    \label{fig:more_depth_vis}
\end{figure*}